\documentclass{article}

\usepackage[preprint,nonatbib]{neurips_2025}

\usepackage[utf8]{inputenc} 
\usepackage[T1]{fontenc}    
\usepackage{hyperref}       
\usepackage{url}            
\usepackage{booktabs}       
\usepackage{amsfonts}       
\usepackage{nicefrac}       
\usepackage{microtype}      
\usepackage{amssymb,amsthm,amsmath}
\usepackage{paralist,hyperref,titlesec,fancyhdr,etoolbox}

\usepackage{comment}
\usepackage{url}
\usepackage{multirow} 
\usepackage[table,xcdraw]{xcolor}
\usepackage{array}
\usepackage{caption}
\usepackage{subcaption} 
\usepackage{pifont}
\usepackage{graphicx} 
\usepackage{tabularx}
\usepackage{wrapfig}
\usepackage[most]{tcolorbox}
\usepackage{listings}
\usepackage{tablefootnote} 
\usepackage{pdflscape}
\usepackage{booktabs}
\usepackage{makecell}
\usepackage{cite}

\usepackage{newunicodechar}
\newunicodechar{−}{\textminus}

\newcolumntype{L}[1]{>{\raggedright\arraybackslash}m{#1}}
\newcolumntype{C}[1]{>{\centering\arraybackslash}m{#1}}
\newcolumntype{R}[1]{>{\raggedleft\arraybackslash}m{#1}}

\newcommand{\figref}[1]{Fig.~\ref{fig:#1}}
\newcommand{\eqnref}[1]{Eq.~(\ref{eqn:#1})}
\newcommand{\tabref}[1]{Table~\ref{tab:#1}}
\newcommand{\secref}[1]{Section~\ref{sec:#1}}
\newcommand{\ssecref}[1]{Section~\ref{subsec:#1}}
\newcommand{\appendixref}[1]{Appendix~\ref{appendix:#1}}

\newcommand{\bhline}[1]{\noalign{\hrule height #1}} 

\newcommand{\shoetsu}[1]{\textcolor{black}{#1}} 
\newcommand{\maru}[1]{\textcolor{black}{#1}} 

\newcommand{\shoetsutwo}[1]{\textcolor{black}{#1}} 
\newcommand{\maruthree}[1]{\textcolor{black}{#1}} 

\title{RATFM: Retrieval-augmented Time Series Foundation Model for Anomaly Detection}

\author{%
  Chihiro Maru \\
  Chuo University\\
  \texttt{cmaru671@g.chuo-u.ac.jp} \\
  \And
  Shoetsu Sato \\
  Independent Researcher\\
  \texttt{jack.and.rozz@gmail.com} \\
}

\begin{document}

\maketitle

\begin{abstract}
Inspired by the success of large language models (LLMs) in natural language processing, recent research has explored the building of time series foundation models and applied them to tasks such as forecasting, classification, and anomaly detection. However, their performances vary between different domains and tasks. In LLM-based approaches, test-time adaptation using example-based prompting has become common, owing to the high cost of retraining. In the context of anomaly detection, which is the focus of this study, providing normal examples from the target domain can also be effective. However, time series foundation models do not naturally acquire the ability to interpret or utilize examples or instructions, because the nature of time series data used during training does not encourage such capabilities. To address this limitation, we propose a retrieval augmented time series foundation model (RATFM), which enables pretrained time series foundation models to incorporate examples of test-time adaptation. We show that RATFM achieves a performance comparable to that of in-domain fine-tuning while avoiding domain-dependent fine-tuning. Experiments on the UCR Anomaly Archive, a multi-domain dataset including nine domains, confirms the effectiveness of the proposed approach.
\end{abstract} 

\section{Introduction}
With the advancement of information technologies, collection of a wide variety of time series data has become increasingly feasible.
In parallel, the importance of anomaly detection has increased, driven by the growing demand for risk control in diverse domains such as health care, industry, and security~\cite{goldberger2000physiobank, johnson2016mimic, hundman2018detecting, su2019robust, lavin2015evaluating}. 
Given the difficulty of obtaining anomalous time series data, many existing methods train models using only in-domain non-anomalous data. 
Specifically, with the progress of deep learning, 
common existing approaches employ
Transformer~\cite{NIPS2017_3f5ee243} to forecast or reconstruct a time series, identifying anomalies based on the degree of deviation from actual observations~\cite{schmidl2022anomaly}.

Inspired by the success of large language models (LLMs) in natural language processing (NLP), there is growing interest in building time series foundation models trained using large-scale time series data~\cite{shi2024timemoe, goswami2024moment, ansari2024chronos, woo2024moirai, das2023decoder}.
These studies have demonstrated significant improvements in many tasks, including time series anomaly detection. They have also reported that model performance can vary across data domains or tasks, and that the fine-tuning of domain-specific data often led to further improvements.
However, it is costly and not feasible to fine-tune a large-scale foundation model to each target domain. 
To address this problem, in-context learning~\cite{DBLP:journals/corr/abs-2005-14165} has attracted increasing attention as a promising alternative in NLP. This approach allows the LLM to adapt to a new domain or task at test time by conditioning on a small number of examples.

This concept is considered particularly effective for anomaly detection because providing a non-anomalous example of the target domain helps a model to distinguish anomalous inputs more easily, even if it does not have prior knowledge of that domain.
However, whether this approach is similarly applicable to time series foundation models remains an open question. Unlike text, which inherently contains cues regarding the domain or task instructions, time series data lacks such explicit semantic structure. As a result, foundation models pretrained by a conventional method might struggle to acquire the capability to interpret examples.

To address this limitation, we propose a retrieval-augmented time series foundation model (RATFM), a framework that enables domain-independent example-based anomaly detection for time series foundation models.
To evaluate the effectiveness of the proposed approach, we conduct experiments on a multi-domain time series dataset, UCR Anomaly Archive~\cite{wu2021current}, using two representative foundation models, Time-MoE~\cite{shi2024timemoe} and Moment~\cite{goswami2024moment}.

Our contributions are summarized as follows:
\begin{itemize}
\item We propose\maruthree{d} RATFM, a method that equip\maruthree{ped} time series foundation models with the ability to leverage a retrieved example. Experiments on a multi-domain dataset demonstrate\maruthree{d} consistent performance improvements across a wide range of domains.
\vspace{-1mm}
\item We propose\maruthree{d} a simple 
post-processing method that significantly improve\maruthree{d} anomaly detection performance. This reveal\maruthree{ed} limitations in the standard scoring method based on simple deviations from the ground truth, and underscore\maruthree{d} the importance of designing an appropriate scoring procedure.
\vspace{-1mm}
\item We perform\maruthree{ed} a detailed analysis of anomaly detection using time series foundation models and reveal\maruthree{ed} unresolved challenges in this task.
\end{itemize}

\section{Related Work}\label{sec:relatedwork}

\noindent \textbf{Time Series Anomaly Detection.} 
Various approaches have been proposed for anomaly detection using time series data, including statistical methods~\cite{yamanishi2002unifying, Sub-PCA}, traditional machine learning techniques~\cite{moshtaghi2014evolving,yeh2016matrix,ding2013anomaly}, and deep learning-based methods~\cite{zamanzadeh2024deep}. 
In particular, deep learning-based methods have recently attracted significant attention due to their ability to model temporal and spatial dependencies. 
In earlier studies, it was common to train models separately for each time series dataset to capture dataset-specific patterns~\cite{liu2024elephant}. 
\shoetsutwo{Research has begun to focus on cross-time series approaches~\cite{zhang2022self, sun2024unraveling, shentu2025towards}, revisiting evaluation metrics~\cite{paparrizos2022volume, 10.1145/3637528.3671971, NEURIPS2018_8f468c87, hwang2019time, 10.1145/3477314.3507024, 10.1145/3534678.3539339}, and the development of more realistic benchmark datasets~\cite{wu2021current, paparrizos2022tsb, Lai2021RevisitingTS}.
However, many models proposed in recent studies are trained in a domain-specific manner, resulting in strong dependency on the dataset~\cite{dai2024sarad, fang2024temporal, feng2024sensitivehue, liu2025gcad, shen2025learn, zhong2025multiresolution, wu2025catch}.
To address this issue, increasing attention is paid to the development of time series foundation models that can be applied across domains and tasks~\cite{shi2024timemoe, goswami2024moment, ansari2024chronos, woo2024moirai, das2023decoder}. This study is closely related to this line of research. We also focus on analyzing anomaly detection results and conducting detailed error analysis, which are often difficult to pursue thoroughly in research that primarily aims to build foundation models.}

\noindent \textbf{Few-shot Examples and Retrieval.} 
\shoetsutwo{Methods that also incorporate examples have been actively studied in recent years in the NLP field. The best-known approach is arguably in-context learning of GPT-3~\cite{DBLP:journals/corr/abs-2005-14165}.
This approach focuses on the use of fixed examples to teach the model the task and expected output format. Our method instead retrieves relevant examples based on the input, making it closer in nature to retrieval-augmented generation (RAG) \cite{DBLP:journals/corr/abs-2005-11401,DBLP:journals/corr/abs-2002-08909}.}
\shoetsutwo{
In the context of anomaly detection and time series analysis, limited efforts have been made to leverage similar examples. Although an existing study employs a fixed set of few-shot examples for multi-modal anomaly detection based on proprietary LLMs~\cite{zhuang2024itthinkitsorted}, such approaches remain largely unexplored to the best of our knowledge.
}



\section{Proposed Method}\label{sec:method}

\subsection{Retrieval-augmented time series foundation model (RATFM)}\label{subsec:FSP}

\begin{figure}[t]
  \begin{minipage}[t]{0.66\linewidth}
    \centering
    \includegraphics[width=\linewidth]{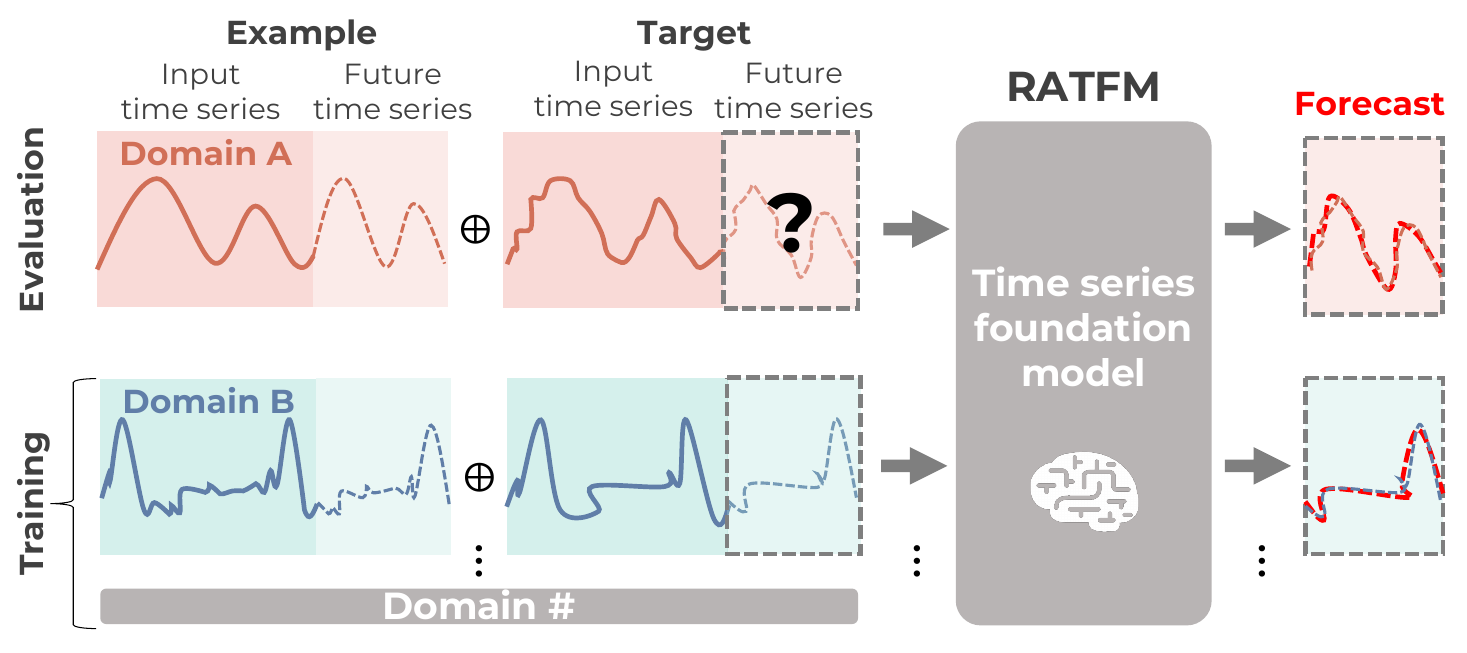}
    \caption{Overview of the retrieval-augmented time series foundation model (RATFM) for forecast-based anomaly detection using time series foundation models.}
    \label{fig:proposed_method}
  \end{minipage}
  \hspace{0.01\linewidth}
  \begin{minipage}[t]{0.33\linewidth}
    \centering
    \includegraphics[width=\linewidth]{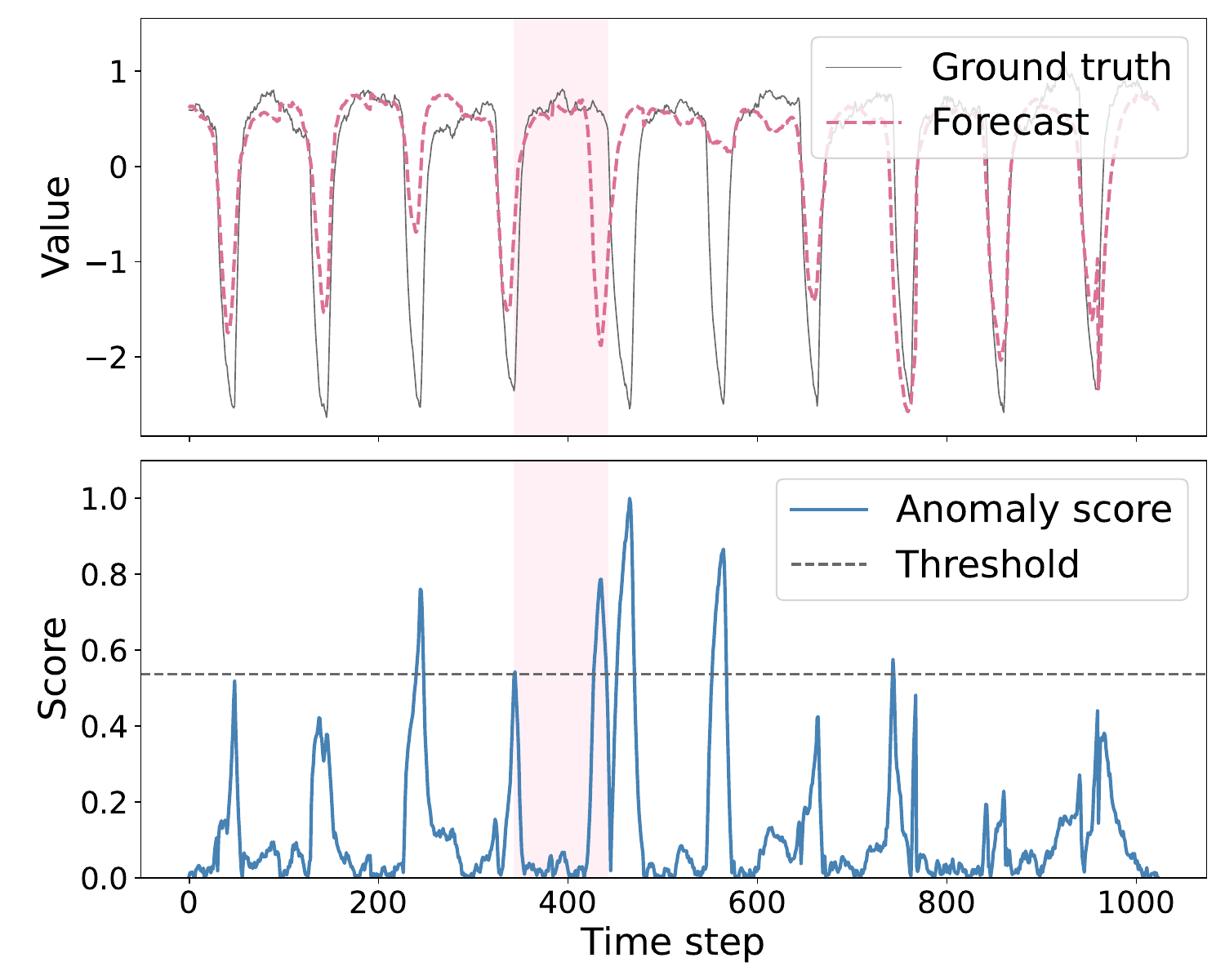}
    \caption{Time series forecasting results (top) and raw anomaly scores (bottom).}
    \label{fig:SMA_1}
  \end{minipage}
\end{figure}

We propose RATFM for forecast-based anomaly detection using time series foundation models, as illustrated in \figref{proposed_method}. RATFM enables domain-independent anomaly detection utilizing retrieved examples.
This capability is valuable in practical settings, \shoetsutwo{where domain-specific fine-tuning is often infeasible owing to model size and training cost.}


\noindent \textbf{\shoetsutwo{Training} with retrieved examples from diverse domains} \quad 
As shown in \figref{proposed_method}, we fine-tune a pretrained time series foundation model. 
Given a target input time series $X^{(d)}_{1:T}$ from domain $d$, the model $f_{\text{forecast}}$ retrieves 
\shoetsutwo{an example that has similar input time series}
$X^{\text{example}(d)}_{1:T}$ from another time series in the same domain, along with its corresponding future time series $X^{\text{example}(d)}_{T+1:T+H}$. To retrieve the example, we
compute
the cross-correlation coefficient, which 
\maruthree{is}
effective and efficient in prior works~\cite{Paparrizos2020DebunkingFL, paparrizos2015k}. The cross-correlation coefficient between two time series $X_{1:T} = \{x_1, \ldots, x_T\}$ and $\tilde{X}_{1:T} = \{\tilde{x}_1, \ldots, \tilde{x}_T\}$ is defined as
\begin{equation}
\label{eqn:similarity}
\frac{CC(X_{1:T}, \tilde{X}_{1:T})}{\|X_{1:T}\| \cdot \|\tilde{X}_{1:T}\|},
\end{equation}
where $CC$ denotes the cross-correlation sequence, and $|\cdot|$ is the Euclidean norm.


\shoetsutwo{The retrieved example is concatenated with the target input time series:}
~$X^{\text{example}(d)}_{1:T} \oplus X^{\text{example}(d)}_{T+1:T+H} \oplus X^{(d)}_{1:T}$. The model is then trained to forecast the future $H$ steps of $X^{(d)}_{1:T}$: 
\begin{equation} 
\hat{X}^{(d)}_{T+1:T+H} = f_{\text{forecast}}\left(X^{\text{example}(d)}_{1:T} \oplus X^{\text{example}(d)}_{T+1:T+H} \oplus X^{(d)}_{1:T}\right). 
\end{equation}
Training 
\maruthree{is}
performed to minimize the mean squared error between the forecast $\hat{X}^{(d)}_{T+1:T+H}$ and ground truth $X^{(d)}_{T+1:T+H}$ across all domains,
\begin{equation} 
\label{eqn:loss_FSP} 
\mathcal{L} = \sum_{d} \left| \hat{X}^{(d)}_{T+1:T+H} - X^{(d)}_{T+1:T+H} \right|^2. 
\end{equation}

\shoetsutwo{Through this training procedure using data from diverse domains, the model acquires a domain-independent ability to leverage retrieved examples as references for time series forecasting.}

\noindent \textbf{Anomaly detection in unseen domains} \quad 
\shoetsutwo{In testing, we assume that anomaly detection is performed on an unseen domain $d^{\prime}$.}
Given a target input time series $X^{(d^{\prime})}_{1:T}$, we similarly 
\maruthree{retrieve}
an example from other time series in the same domain, and forecast $\hat{X}^{(d^{\prime})}_{T+1:T+H}$ 
using the model which RATFM 
\maruthree{is}
applied to.
The anomaly scores for each data point $x_{t}^{(d^{\prime})}$ in $X^{(d^{\prime})}_{T+1:T+H}$ are computed as

\begin{equation} 
\label{eqn:normal_anomaly_score}
\begin{aligned} 
AS(x_{t}^{(d^{\prime})}) = \left|\hat{x}_t^{(d^{\prime})} - x_t^{(d^{\prime})}\right|. 
\end{aligned} 
\end{equation}

Data points of which anomaly scores exceed a predefined threshold 
are identified as anomalies.

\shoetsutwo{The problem for which our method is particularly effective is the occurrence of false positive anomalies in unseen domains. Typically, forecast-based anomaly detection is designed under the assumption that forecasts succeed for non-anomalous data and fail for anomalous data, resulting in high anomaly scores only for the latter.
However, in unseen domain anomaly detection without retrieved examples, forecasts often fail even for non-anomalous data, leading to elevated anomaly scores and reduced precision. Furthermore, because threshold-based anomaly scoring relies on the overall distribution of anomaly scores, such falsely high scores can raise the threshold, potentially causing subtle anomalies to be overlooked.}

\subsection{Anomaly Score Smoothing via Simple Moving Average}\label{subsec:SMA}
\shoetsutwo{We also propose a method for computing anomaly scores, addressing issues we identified through an analysis of existing approaches.}
In many anomaly detection models, the anomaly scores often do not match the anomalies perceived by humans.
As a preliminary experiment, we examined the time series forecasting results and the corresponding anomaly scores generated by a baseline model, as shown in \figref{SMA_1}.\footnote{Here, we employed \textbf{Time-MoE} with the \textbf{Zero-shot} setting, as detailed in \secref{settings}.} 
\shoetsutwo{We observed that the anomaly scores tend to increase significantly at periodic peaks, regardless of whether true anomalies are present. This is because the anomaly score is computed based on the degree of deviation from the ground truth, which tends to be high at such peaks. 
Consequently, these elevated scores in the peak regions can obscure the true anomalies. 
}

\shoetsutwo{To mitigate the effect of peaks,} we propose a post-processing method that applies a simple moving average (SMA)~\cite{chatfield2004timeseries} to the raw anomaly scores. SMA is a commonly used smoothing method that computes the average over a fixed-length window. Given a data point $x_t$, the smoothed anomaly score $AS(x_t)$ using an $n$-point moving average is defined as follows:
\begin{equation}
\label{eqn:SMA_anomaly_score}
\begin{aligned}
 AS(x_{t}) = \frac{1}{n} \sum_{i=0}^{n-1} x_{t-i}.
\end{aligned}
\end{equation}

In our experiments, $n$ was set as the estimated period of each time series obtained using the Fourier transform~\cite{chatfield2004timeseries}. 

\section{Experimental Settings}\label{sec:settings}

\subsection{Datasets}\label{subsec:dataset}

\shoetsutwo{To confirm that our proposed method enables models to perform anomaly detection in unseen domains,}
we conducted experiments using the UCR Anomaly Archive~\cite{wu2021current}, which consists of 250 univariate time series of nine different domains.
The statistics 
of the dataset are included in \appendixref{datasets}.
Each time series was divided into training and testing data, and each testing data contained an anomaly.
The position and duration of the anomaly varied across the time series.
As a preprocessing step, we applied standardization
to normalize the time series to a common scale~\cite{Z-score}. Specifically, each data point $x_t$ in both the training and test data was standardized using the mean and standard deviation of the training data. 




\subsection{Models}
\smallskip\noindent\textbf{Time-MoE}\footnote{\url{https://github.com/Time-MoE/Time-MoE}}~\cite{shi2024timemoe}:
\shoetsutwo{We employed Time-MoE (large), which is a decoder-only time series foundation model with a mixture-of-experts architecture for time series forecasting.}

\smallskip\noindent\textbf{Moment}\footnote{\url{https://github.com/moment-timeseries-foundation-model/moment}}~\cite{goswami2024moment}:
\shoetsutwo{Moment is a time series foundation model trained through reconstruction tasks. 
It can be adapted to various downstream tasks by modifying and fine-tuning its lightweight linear head. 
Although we basically employed this model as a reconstruction-based anomaly detector, we evaluated RATFM under both reconstruction and forecasting settings because recent anomaly detection methods increasingly employ forecast-based approaches. 
}

\smallskip\noindent\textbf{Anomaly Transformer}\footnote{\url{https://github.com/thuml/Anomaly-Transformer}}~\cite{xu2022anomaly}\shoetsu{:}
Anomaly Transformer is a \shoetsutwo{neural-based anomaly detection model that reconstructs input time series and computes anomaly scores.} 

\smallskip\noindent\textbf{Sub-PCA}\footnote{\url{https://github.com/TheDatumOrg/TSB-AD}}~\cite{Sub-PCA}\shoetsu{:}
Sub-PCA is a reconstruction-based statistical method that leverages principal component analysis. We selected this method because it demonstrated strong performance in a comparative study of various anomaly detection techniques, including neural-based methods~\cite{liu2024elephant}. 

\smallskip\noindent\textbf{GPT-4o}\footnote{\url{https://github.com/Rose-STL-Lab/AnomLLM}}~\cite{openai2024gpt4technicalreport}:
\shoetsutwo{
We also evaluated a representative text-based foundation model, GPT-4o.
Following a prior work~\cite{zhou2025can}, we input a time series as a text with a prompt asking to detect an anomaly span. As this method does not output continuous anomaly scores, we only report point-wise F1 scores. The detailed prompt is described in \appendixref{GPTprompt}.
}



\subsection{Training Settings}

The following experimental settings were used for \textbf{Time-MoE} and \textbf{Moment}.

\smallskip\noindent\textbf{Zero-shot}:
The pretrained model was used without additional training or adaptation. 

\smallskip\noindent\textbf{Out-domain fine-tuning (FT)}:
\shoetsutwo{The pretrained model was fine-tuned from out-domain data (i.e., the eight non-target domains from the UCR Anomaly Archive for each target domain). }


\smallskip\noindent\textbf{In-domain fine-tuning (FT)}:

\begin{wrapfigure}{r}{0.5\textwidth} 
  \centering
  \vspace{-1em}  
  \includegraphics[width=0.48\textwidth]{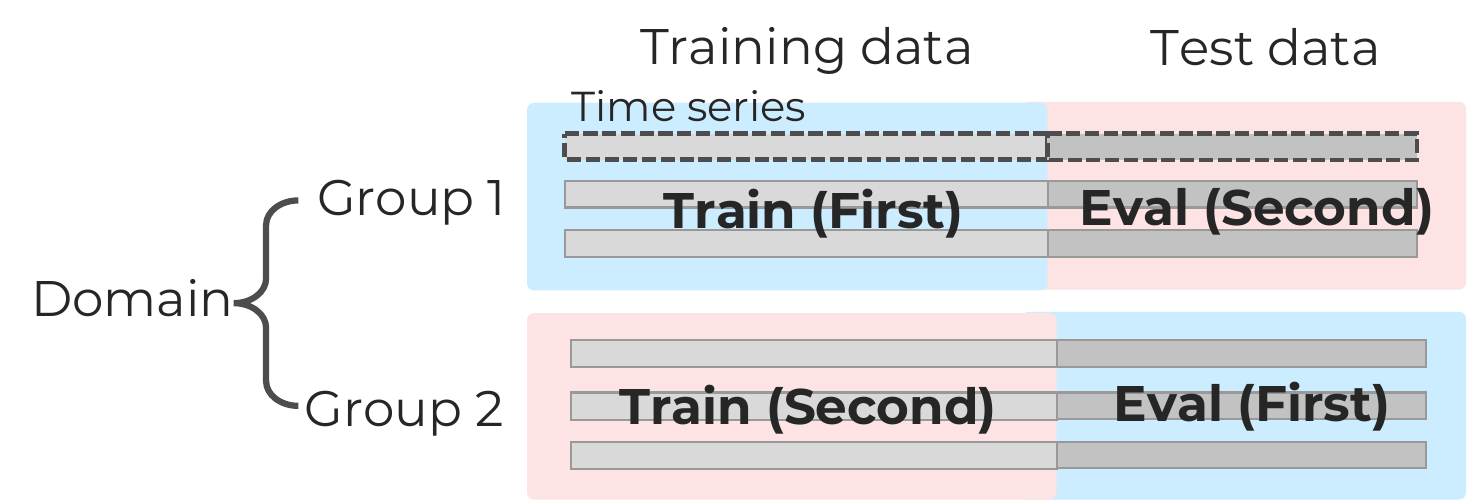}
  \caption{Training and test split for \textbf{In-domain FT}.}
  \label{fig:setting_in-domain_FT}
\end{wrapfigure}

The pretrained model was fine-tuned from the target domain data. Specifically, the time series in each domain were divided into two groups, as illustrated in \figref{setting_in-domain_FT}. The model was fine-tuned on one group and anomaly detection was performed on the other.\footnote{In many anomaly detection datasets, training and test data are generated by splitting a single time series.
As a result, 
\textbf{In-domain FT} can introduce two types of bias: (1) a bias due to the same domain and (2) a bias caused by training data coming from the same time series as the test data.
To remove the latter, we adopted a two-fold split strategy, in which the training and test data came from different time series.}
The roles of the two groups were then exchanged. 
\shoetsutwo{Note that we regard this setting as a costly upper bound.}

\smallskip\noindent\textbf{RATFM}:
\shoetsutwo{We retrained the baseline model from the same data as \textbf{Out-domain FT},
following the procedure 
in \ssecref{FSP}.\footnote{Note that this does not mean a need for domain-specific fine-tuning, as we are merely simulating an out-domain scenario in this experiment.}}

\smallskip\noindent\textbf{RATFM w/o training}:
\shoetsutwo{The same input as that of \textbf{RATFM} was fed to a \textbf{Zero-shot} model. This setting aimed to examine the capability of vanilla time series foundation models to use examples.}

Because the existing methods do not perform fine-tuning, we applied the following common experimental settings to \textbf{Anomaly Transformer} and \textbf{Sub-PCA}.

\smallskip\noindent\textbf{Time Series-wise}: 
This setting, which has been used in many previous studies on anomaly detection, trains a separate model for each time series, even when they belong to the same domain.

\smallskip\noindent\textbf{In-domain}: 
We conduct\maruthree{ed} model training and evaluation under the same experimental setting for a fair comparison with \textbf{In-domain FT}.

\tabref{setting_length} summarizes how the input length was allocated for each training setting.\footnote{We ran the experiments on a workstation equipped with a 48-core AMD EPYC CPU, an NVIDIA A100 GPU, and 256GB of RAM.}
Because the raw time series in the dataset were typically much longer than the maximum input length supported by the models, we segmented the data accordingly. Following prior work~\cite{shi2024timemoe}, we set the forecast horizon to 96 and adjusted the input depending on the model architecture and training strategy. We set the 
input length to be identical for each model to ensure a fair comparison across different settings. For example, in \textbf{Time-MoE} with \textbf{RATFM}, the maximum input length of 1120 was split into 512 for the example input, 96 for the example future time series, and 512 for the target input. In contrast, in \textbf{Zero-shot}, where no examples are used, the entire input length (1120) was allocated to the input sequence.

\begin{table}[t]
\footnotesize
\caption{Experimental configurations by training setting.}
\label{tab:setting_length}
\resizebox{\textwidth}{!}{%
\begin{tabular}{llll}
\bhline{1.25pt}
\multicolumn{1}{c}{Base Model} & \multicolumn{1}{c}{Training Setting} & \multicolumn{1}{c}{Input Length} & \multicolumn{1}{c}{Parameters} \\
\hline
\multirow{2}{*}{\textbf{Time-MoE}} & \textbf{RATFM} & $1120 = (512+96)+512$ & \multirow{2}{*}{453M} \\
                                   & Other settings & $1120$                 & \\
\hline
\multirow{3}{*}{\textbf{Moment}}   & \textbf{RATFM (forecast)} & $512 = (208+96)+208$ & \multirow{3}{*}{385M} \\
                                   & \textbf{RATFM (reconstruction)} & $512 = (160+96)+(160+96)$ & \\
                                   & Other settings & $512$                  & \\
\hline
\textbf{Anomaly Transformer} & All settings & $512$ & 4.75M \\
\hline
\textbf{Sub-PCA} & All settings & $512$ & -- \\
\hline
\textbf{GPT-4o} & \textbf{Zero-shot} & $96$ & undisclosed\tablefootnote{Considering that recent open LLMs range from several billion to hundreds of billions of parameters, we assume that it is at least 100 times larger than other models we employed.} \\
\bhline{1.25pt}
\end{tabular}
}
\end{table}

\subsection{Evaluation Metrics} \label{ssec:evaluation_metric}
We used the point-wise F1 score, VUS-ROC, and VUS-PR~\cite{paparrizos2022volume} as evaluation metrics. 
\shoetsutwo{To compute the F1 score, we binarized the anomaly score of each data point based on a threshold, which was defined as the mean plus three standard deviations of the anomaly scores in the test data following a prior work~\cite{paparrizos2022volume}.}
\shoetsutwo{A well-known limitation of point-wise metrics is their sensitivity to slight misalignments in detection timing; even minor discrepancies between the forecast and ground truth labels can result in penalties. VUS-ROC and VUS-PR are threshold-free metrics used to handle the problem where ground truth anomaly labels are treated as continuous values.
}

\shoetsutwo{We did not adopt the adjusted-F1 score~\cite{xu2018unsupervised}, a modified version of the point-wise F1 score that has been frequently used in prior work. This metric regards an anomaly segment as successfully detected if the model labels at least one point within the segment as anomalous correctly. Although widely used, it often results in unrealistically high recall and hinders fair comparison~\cite{liu2024elephant}.}


\section{Experimental Results}\label{sec:results}

\subsection{Effect of Simple Moving Average}



\begin{table}[!t]
\footnotesize
\caption{Effect of SMA on VUS-ROC performance.}
\label{tab:SMA-results}
\centering
\begin{tabular}{l|cr|cr}
\bhline{1.25pt}
\multirow{2}{*}{VUS-ROC} & \multicolumn{2}{c|}{\textbf{Zero-shot}} & \multicolumn{2}{c}{\textbf{RATFM}} \\
 & \textbf{w/o SMA} & \multicolumn{1}{c|}{\textbf{w/ SMA}} & \textbf{w/o SMA} & \multicolumn{1}{c}{\textbf{w/ SMA}} \\ \hline
\textbf{Time-MoE} & \multicolumn{1}{r}{65.7\%} & \textbf{68.7\%} & \multicolumn{1}{r}{68.9\%} & \textbf{76.1\%} \\
\textbf{Moment} & \multicolumn{1}{r}{64.9\%} & \textbf{70.6\%} & \multicolumn{1}{r}{68.8\%} & \textbf{74.3\%} \\ \bhline{1.25pt}
\end{tabular}
\end{table}

\begin{figure}[!t]
\centering

\begin{subfigure}{0.33\columnwidth}
    \centering
    \includegraphics[width=\linewidth]{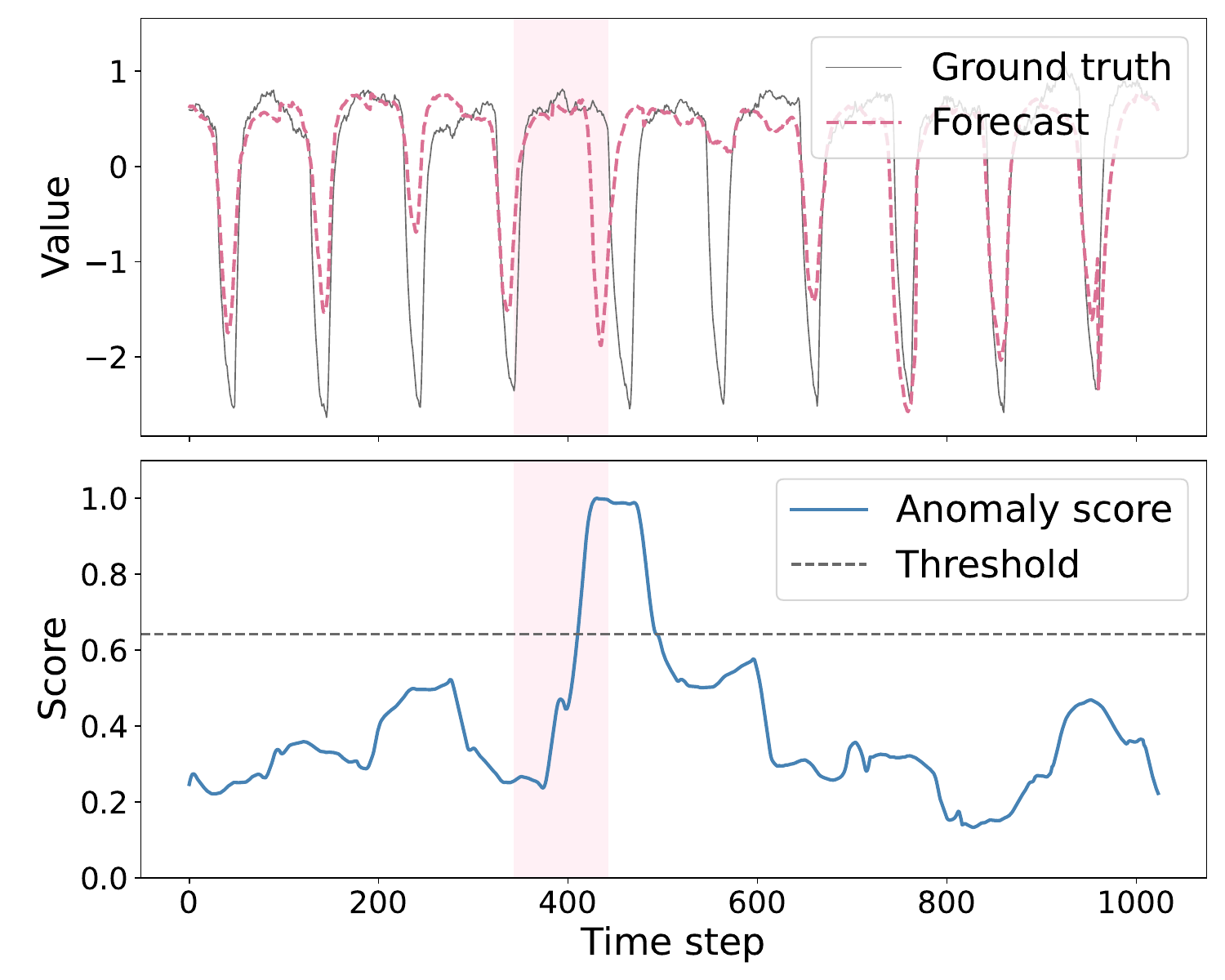}
    \caption{\textbf{Zero-shot}.}
    \label{fig:SMA-zero_shot_anomalous}
\end{subfigure}
\begin{subfigure}{0.33\columnwidth}
    \centering
    \includegraphics[width=\linewidth]{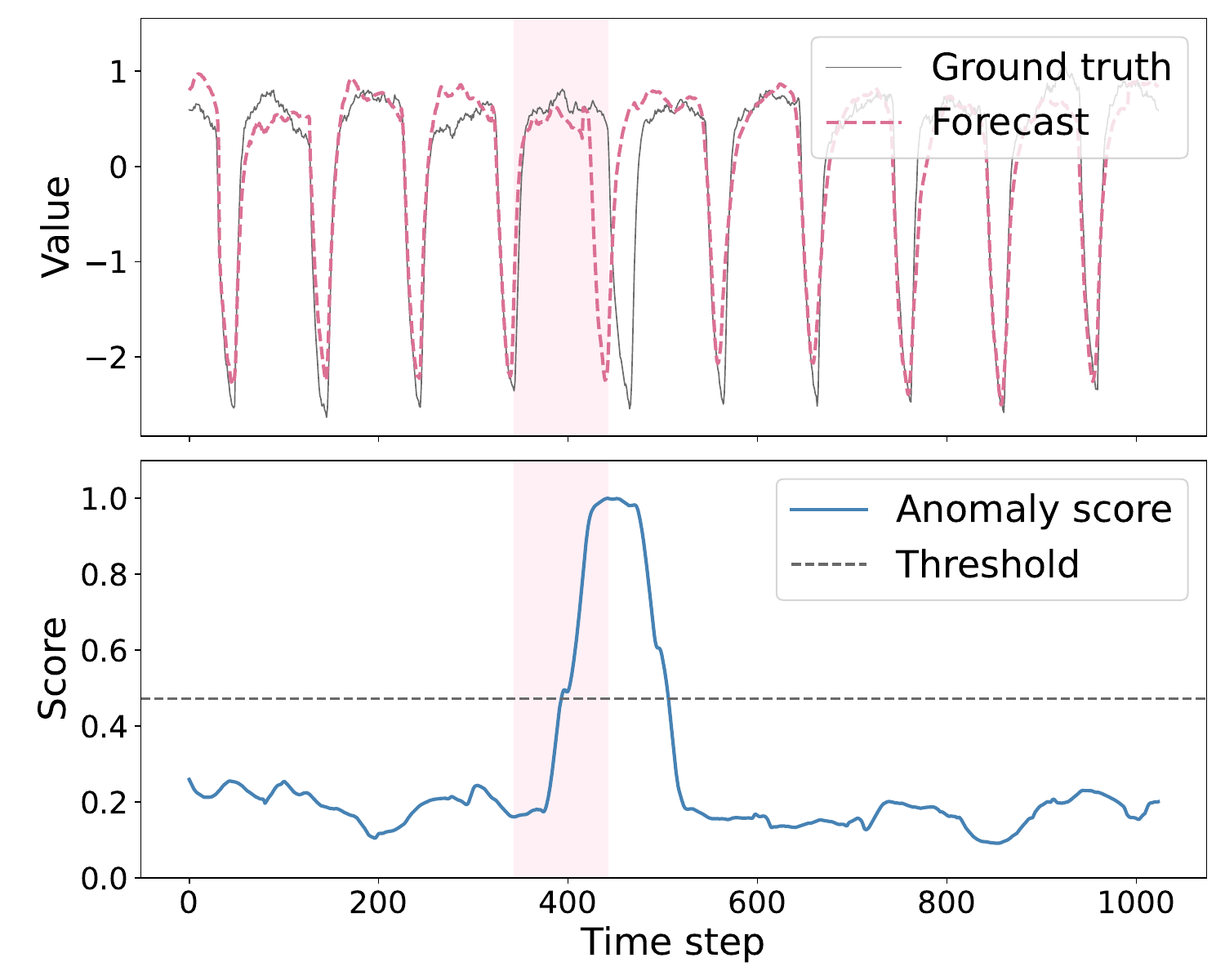}
    \caption{\textbf{RATFM}.}
    \label{fig:SMA-few-shot_prompting_anomalous}
\end{subfigure}

\caption{Anomaly scores 
of \textbf{Time-MoE}
after applying SMA. The time series is the same as that in \figref{SMA_1}. The highlighted regions indicate the ground-truth anomalies.}
\label{fig:SMA_2}
\end{figure}
\shoetsutwo{We first evaluated the proposed SMA-based post-processing, as the choice of scoring method influences the subsequent experimental results.}
We compared two approaches:
1) using the raw anomaly scores without any post-processing, as defined in \eqnref{normal_anomaly_score} (\textbf{w/o SMA}), and
2) applying SMA described in \eqnref{SMA_anomaly_score} to smooth the raw anomaly scores (\textbf{w/ SMA}).

As presented in \shoetsutwo{\tabref{SMA-results}, the application of SMA significantly improved VUS-ROC in all settings.}
\figref{SMA_2} also visualizes the results of the time series forecast and the corresponding anomaly scores after applying SMA. The time series is the same as that in \figref{SMA_1}.
Although \figref{SMA_1} shows high anomaly scores at almost every periodic peak, these false positives
are suppressed in \figref{SMA_2}. \shoetsutwo{Based on these results, we report the performance after applying SMA in the subsequent experiments.
}

\subsection{Performance Evaluation}\label{subsec:main-results-vusroc}

\begin{table}[t]
\footnotesize
\centering
\caption{Comparison of model performance after applying SMA.}
\label{tab:main-results-vusroc}
\resizebox{\textwidth}{!}{
\begin{tabular}{llrrrrr}
\bhline{1.25pt}
\multicolumn{1}{c}{\multirow{2}{*}{Base Model}} & \multicolumn{1}{c}{\multirow{2}{*}{Training Setting}} & \multicolumn{1}{c}{\multirow{2}{*}{VUS-ROC}} & \multicolumn{1}{c}{\multirow{2}{*}{VUS-PR}} & \multicolumn{3}{c}{Point-wise} \\
\multicolumn{1}{c}{} & \multicolumn{1}{c}{} & \multicolumn{1}{c}{} & \multicolumn{1}{c}{} & \multicolumn{1}{c}{F1 Score} & \multicolumn{1}{c}{Precision} & \multicolumn{1}{c}{Recall} \\ \hline
\multirow{5}{*}{\textbf{Time-MoE}} 
 & \textbf{Zero-shot} & 68.7\% & 14.7\% & 11.3\% & 12.7\% & 13.2\% \\
 & \textbf{Out-domain FT} & 70.5\% & 13.9\% & 9.7\% & 10.4\% & 13.1\% \\
 & \textbf{In-domain FT} & 79.1\% & 20.5\% & 16.1\% & 14.8\% & 23.8\% \\
 & \textbf{RATFM} & 76.1\% & 17.7\% & 13.2\% & 12.9\% & 17.7\% \\ 
 & \textbf{RATFM w/o training} & 65.9\% & 15.7\% & 11.8\% & 13.5\% & 13.4\% \\ \hline
\multirow{6}{*}{\textbf{Moment}} 
 & \textbf{Zero-shot} & 70.6\% & 15.1\% & 9.9\% & 10.9\% & 15.2\% \\
 & \textbf{Out-domain FT} & 70.7\% & 16.1\% & 11.4\% & 13.3\% & 16.9\% \\
 & \textbf{In-domain FT} & 77.4\% & 21.7\% & 14.3\% & 12.8\% & 27.2\% \\
 & \textbf{RATFM (forecast)} & 74.3\% & 16.3\% & 12.6\% & 13.2\% & 16.1\% \\
 & \textbf{RATFM w/o training} & 69.0\% & 16.2\% & 9.7\% & 13.0\% & 13.4\% \\ 
 & \textbf{RATFM (reconstruction)} & 67.1\% & 15.8\% & 10.0\% & 10.5\% & 17.3\% \\ \hline
\multirow{2}{*}{\textbf{Anomaly Transformer}} 
 & \textbf{Time Series-wise} & 51.8\% & 2.9\% & 1.2\% & 1.2\% & 2.4\% \\
 & \textbf{In-domain} & 53.5\% & 4.0\% & 2.4\% & 2.3\% & 5.3\% \\ \hline
\multirow{2}{*}{\textbf{Sub-PCA}} 
 & \textbf{Time Series-wise} & 64.9\% & 11.3\% & 7.8\% & 8.6\% & 10.1\% \\
 & \textbf{In-domain} & 61.3\% & 9.3\% & 6.6\% & 7.6\% & 8.4\% \\ \hline
\textbf{GPT-4o} & \textbf{Zero-shot} & -- & -- & 1.5\% & 1.0\% & 8.9\% \\ \bhline{1.25pt}
\end{tabular}
}
\end{table}


\begin{figure*}[t]
    \centering
    \begin{subfigure}{0.33\textwidth} 
        \centering
        \includegraphics[width=\textwidth]{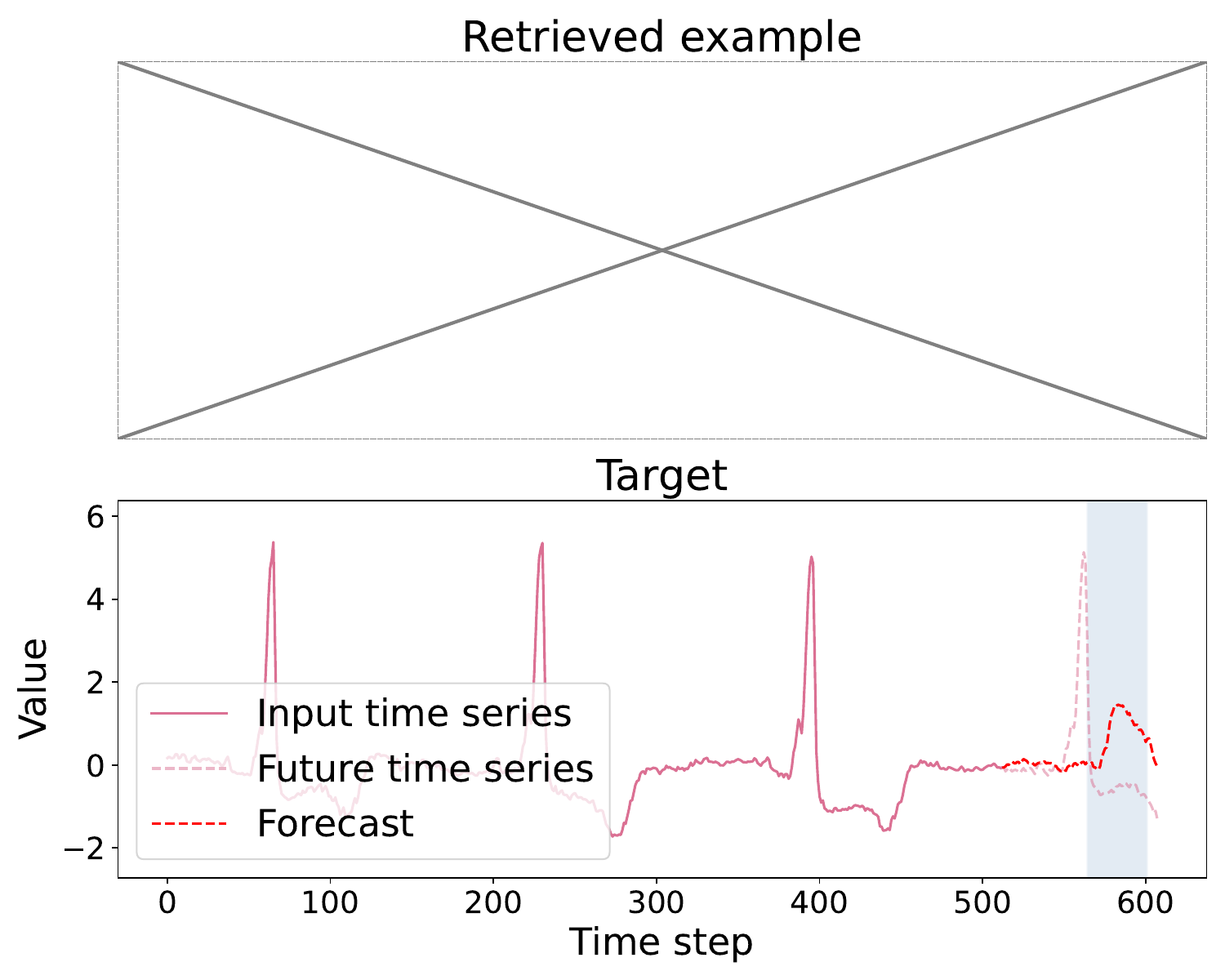}
        \caption{\textbf{Out-domain FT}.}
    \end{subfigure}
    \hspace{-0.5em} 
    \begin{subfigure}{0.33\textwidth} 
        \centering
        \includegraphics[width=\textwidth]{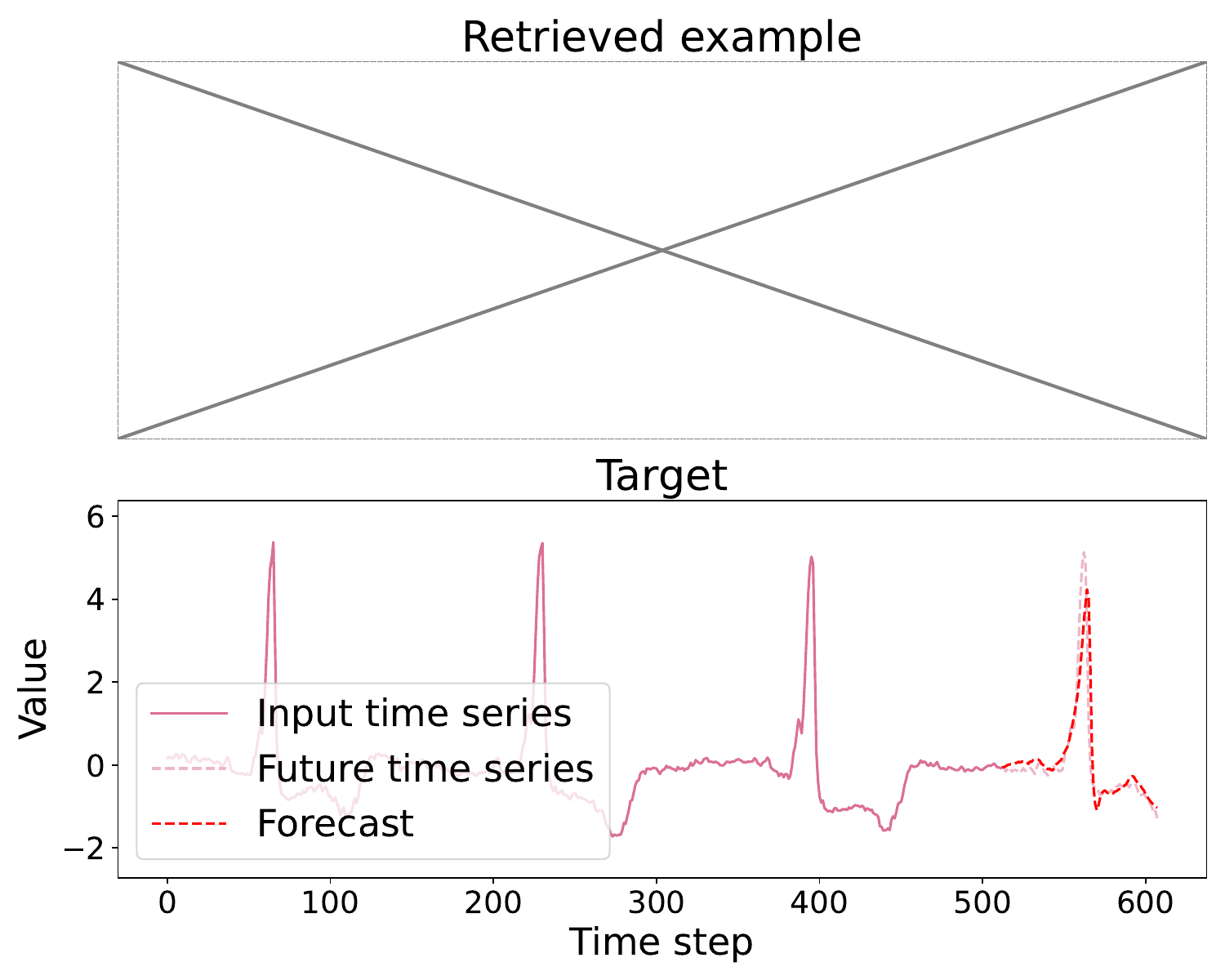}
        \caption{\textbf{In-domain FT}.}
    \end{subfigure}
    \hspace{-0.5em} 
    \begin{subfigure}{0.33\textwidth}
        \centering
        \includegraphics[width=\textwidth]{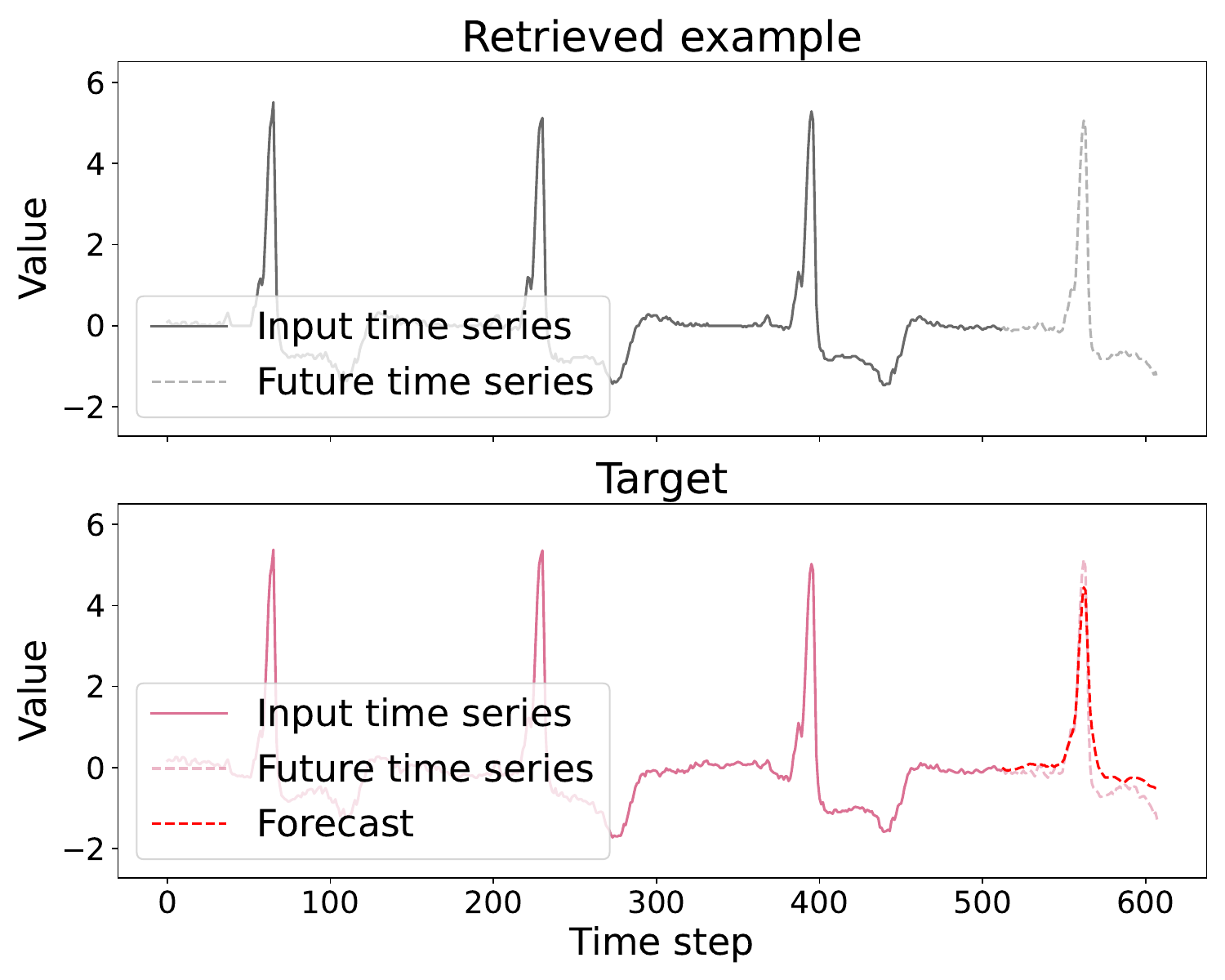}
        \caption{\textbf{RATFM}.}
    \end{subfigure}
    \caption{Target (bottom) and retrieved example (top) in \textbf{Time-MoE}. The blue-highlighted regions indicate false positive anomalies.}
    \label{fig:best_match_examples}
\end{figure*}


\tabref{main-results-vusroc} presents the main results, including a comparison of all models.\footnote{The results computed from raw anomaly scores are also shown in \tabref{main-results-vusroc-noSMA}.}
First, both \textbf{Time-MoE} and \textbf{Moment} with \textbf{RATFM} 
outperformed \textbf{Zero-shot} and \textbf{Out-domain FT} in all metrics.
Note that \textbf{RATFM} also achieved comparable performance to that of \textbf{In-domain FT} although it does not require domain-dependent training. 
Specifically, \figref{best_match_examples} shows the results in different settings for the same time series; although \textbf{Out-domain FT} sometimes failed to forecast and had false positive regions (blue), \textbf{RATFM} stably forecast the time series by referring a similar example.

\shoetsutwo{As expected, \textbf{In-domain FT} achieved the best performance because it had domain-specific knowledge. 
Even in the \textbf{Zero-shot} setting, the time series foundation models outperformed other existing methods, demonstrating its strong generalization capability.}
\textbf{Out-domain FT}, which was fine-tuned with the same out-domain data as \textbf{RATFM}, showed no substantial improvement compared with the \textbf{Zero-shot} setting. These results also confirmed
the domain-independent anomaly detection capability of \textbf{RATFM}.
These findings were consistent with the domain-wise results listed in \tabref{main-result-domain}.
\shoetsutwo{
Interestingly, \textbf{RATFM} with reconstruction-based \textbf{Moment} did not improve. Due to space limitations, we present a detailed discussion in \appendixref{forecasting_vs_reconstruction}.}
Although ChatGPT demonstrates strong performance in a wide range of tasks, \textbf{GPT-4o} underperformed in anomaly detection compared to time series foundation models. A possible explanation is that \textbf{GPT-4o} struggled to capture meaningful patterns in numerical sequences when time series \maruthree{were} provided as text via prompts.

\section{Analysis}\label{ssec:analysis}
\begin{table}[t]
\centering
\footnotesize
\caption{Average similarity between the forecast target and the input under each setting of \textbf{Time-MoE}.}
\label{tab:similarity_average}
\begin{tabular}{l rr}  
\bhline{1.25pt}
\multicolumn{1}{c}{Setting} & 
\multicolumn{1}{c}{Lengths} & 
\multicolumn{1}{r}{Avg. Similarity} \\

\hline
(a) \textbf{RATFM} (example future time series and forecast target)  & $(96, 96)$  & $0.974$ \\
(b) \textbf{Zero-shot} (partial input time series and forecast target)   & $(96, 96)$ & $-0.015$ \\
(c) \textbf{Zero-shot} (partial input time series and forecast target)  & $(608, 96)$ & $0.768$ \\
\bhline{1.25pt}
\end{tabular}
\end{table}

\subsection{\shoetsutwo{Why Are Retrieved Examples More Effective Than Input Time Series?}}

\begin{figure}[t]
    \centering
    \begin{minipage}[t]{0.37\linewidth}
        \centering
        \includegraphics[width=\linewidth]{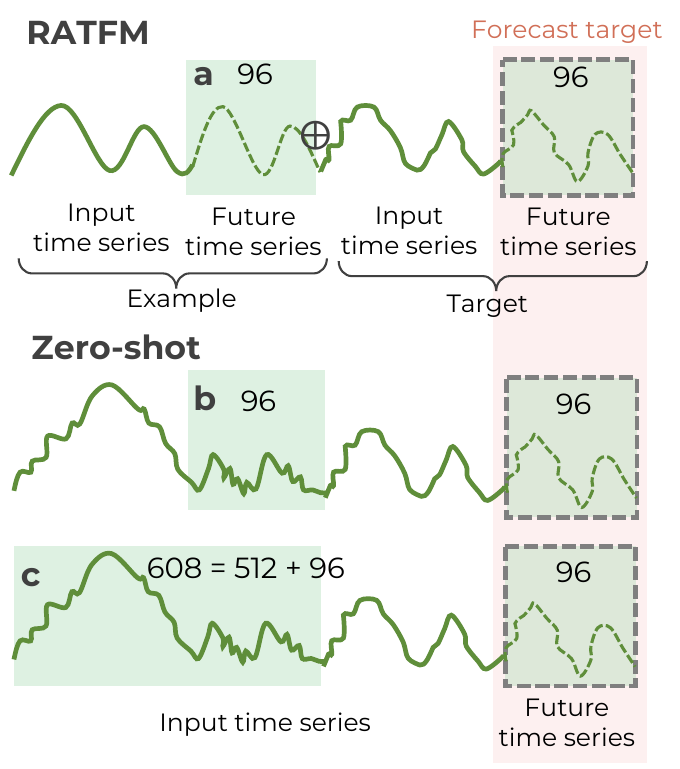}
        \caption{Segments used for computing similarity with the forecast target.}
        \label{fig:similarity}
    \end{minipage}
    \hspace{0.01\linewidth}
    \begin{minipage}[t]{0.6\linewidth}
        \centering
        \includegraphics[width=\linewidth]{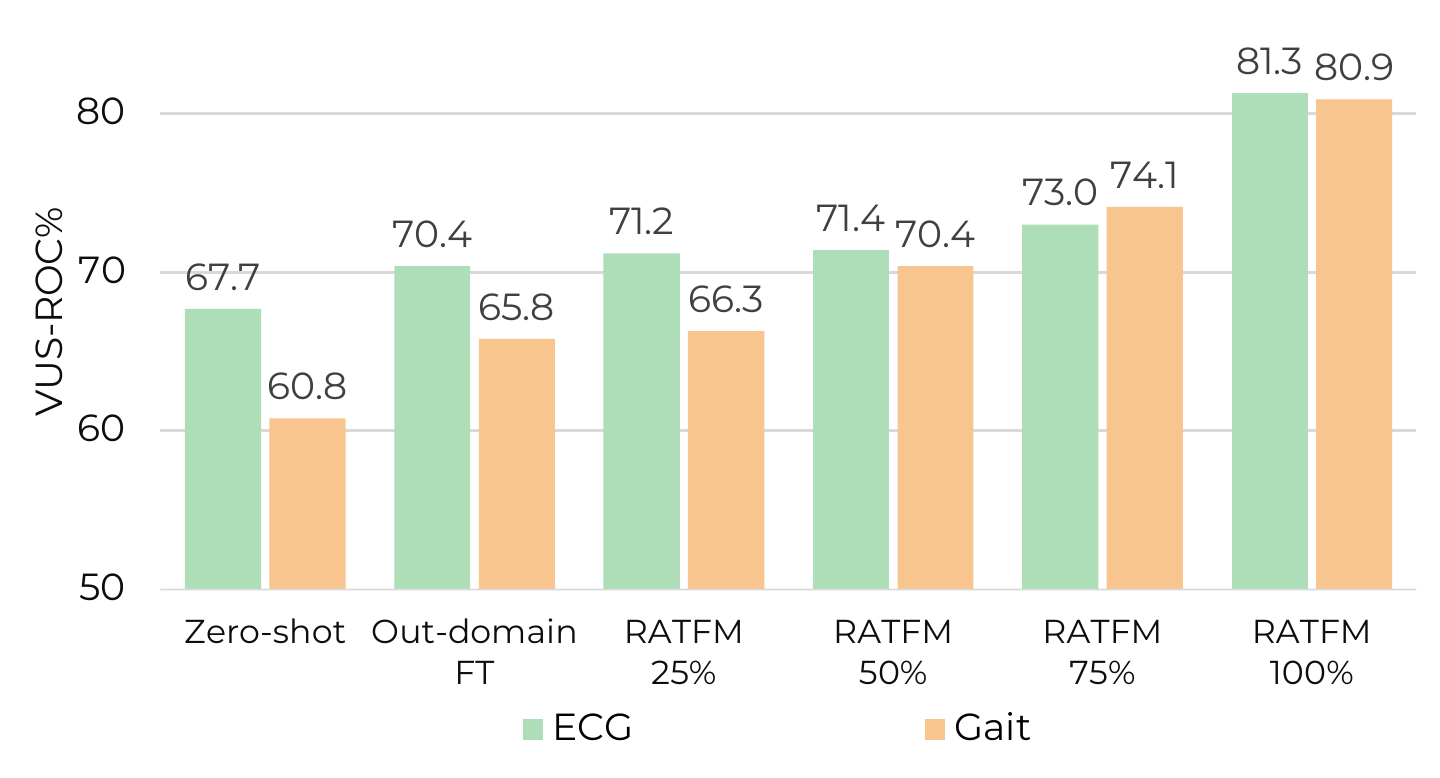}
        \caption{Effect of the number of candidate examples on VUS-ROC performance.}
        \label{fig:candidate_example}
    \end{minipage}%
\end{figure}



\shoetsutwo{Although \textbf{RATFM} consistently outperformed \textbf{Zero-shot} as discussed in \ssecref{main-results-vusroc}, a reasonable question remains: time series data for anomaly detection is often cyclical --- so why did the \textbf{Zero-shot} setting fail to leverage the input itself as a similar example?}
To address the question,
\tabref{similarity_average} and \figref{similarity} present
the similarities
between the forecast target and three types of time series segments:
\textbf{(a)} the example future time series in \textbf{RATFM};
\textbf{(b)} the segment of \textbf{Zero-shot} that corresponds to the example future time series in \textbf{RATFM};
and
\textbf{(c)} the segment of \textbf{Zero-shot} that corresponds to the entire example time series in \textbf{RATFM}.

\shoetsutwo{
These results imply two tendencies: 1) a retrieved example tends to be more similar to the forecast target than the input itself (0.974 vs 0.768), and 2) even if the input time series is cyclical, segments similar to the forecast target may not appear at precisely aligned positions (−0.015 vs 0.768).\footnote{Domain-wise similarities are shown in \tabref{similarity_domain}.}
Moreover, since the time series foundation model is also trained on non-cyclical data, it likely has not acquired the ability to utilize a cyclical input as examples during pretraining, as suggested by the low performance of \textbf{RATFM w/o training} in \tabref{main-results-vusroc}.
This implies that the ability to leverage examples effectively needs to be explicitly learned through post-hoc training like \textbf{RATFM}.
}

\subsection{Does RATFM Still Work with Fewer Example Candidates?}
We assess\maruthree{ed} the impact of reducing the number of candidate time series used for example retrieval in \textbf{RATFM}, since it can be difficult to collect many in-domain candidates in practice.
\figref{candidate_example} shows the VUS-ROC for the electrocardiogram (ECG) and gait signal (Gait) domains in the dataset.
We reduced the size of the candidate pool to 100\%, 75\%, 50\%, and 25\% of its original size.
Although the VUS-ROC generally decrease\maruthree{d} as the number of candidate examples \maruthree{was} reduced, the performance remain\maruthree{ed} above the \textbf{Zero-shot} and \textbf{Out-domain FT} baselines, even at 25\%.
These results indicate that \textbf{RATFM} maintains strong performance even when candidate availability is limited, demonstrating its practicality in situations where collecting a large number of examples is challenging.


\subsection{Error Analysis}
\label{subsec:error-analysis}

\begin{figure}[t]
    \centering
    \begin{subfigure}{0.34\textwidth} 
        \centering
        \includegraphics[width=\textwidth]{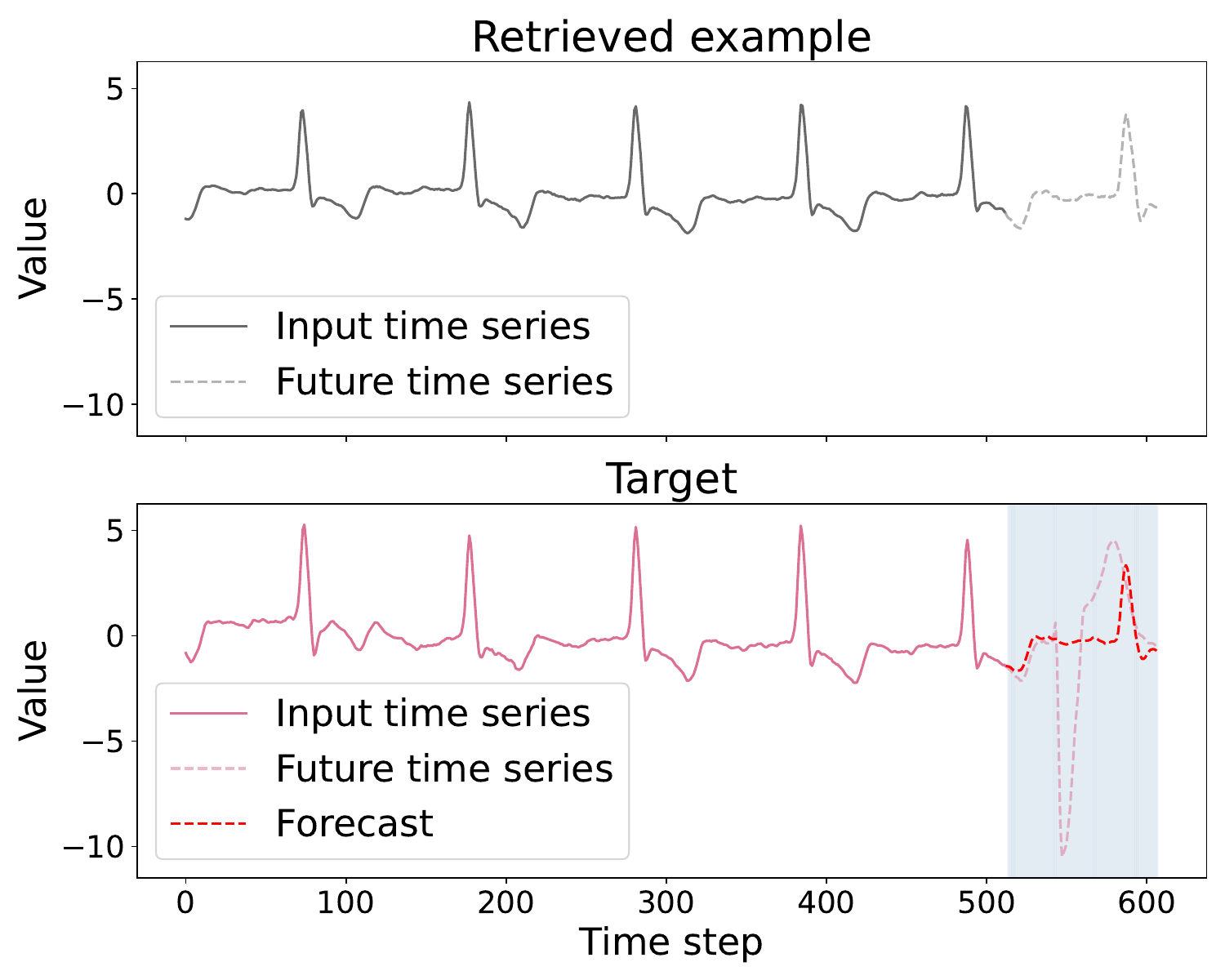}
        \caption{Dissimilar future time series between target and example.}
        \label{fig:subfig_a}
    \end{subfigure}
    \begin{subfigure}{0.32\textwidth}
        \centering
        \includegraphics[width=\textwidth]{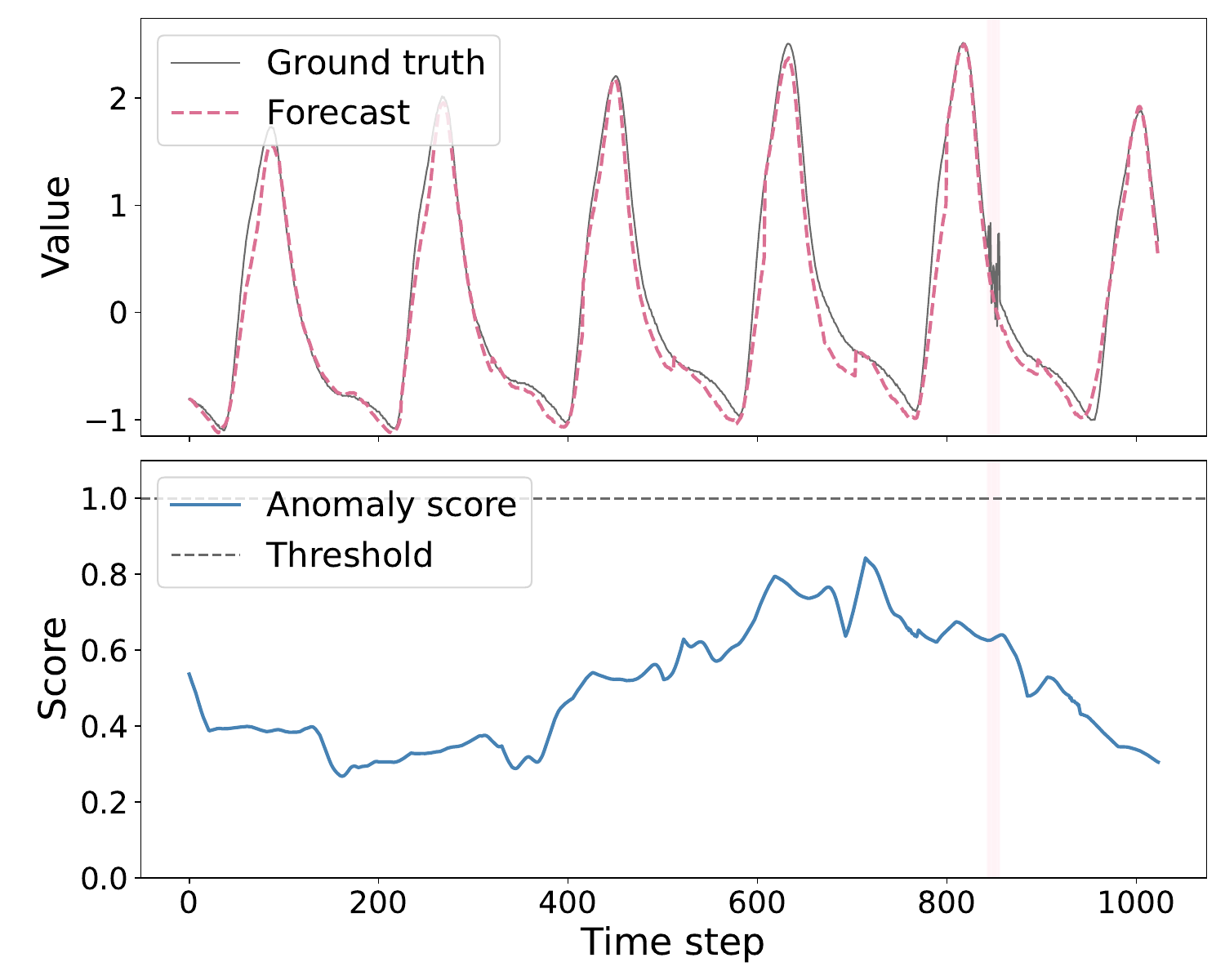}
        \caption{Small deviations from normal patterns.}
        \label{fig:subfig_b}
    \end{subfigure}
    \begin{subfigure}{0.32\textwidth}
        \centering
        \includegraphics[width=\textwidth]{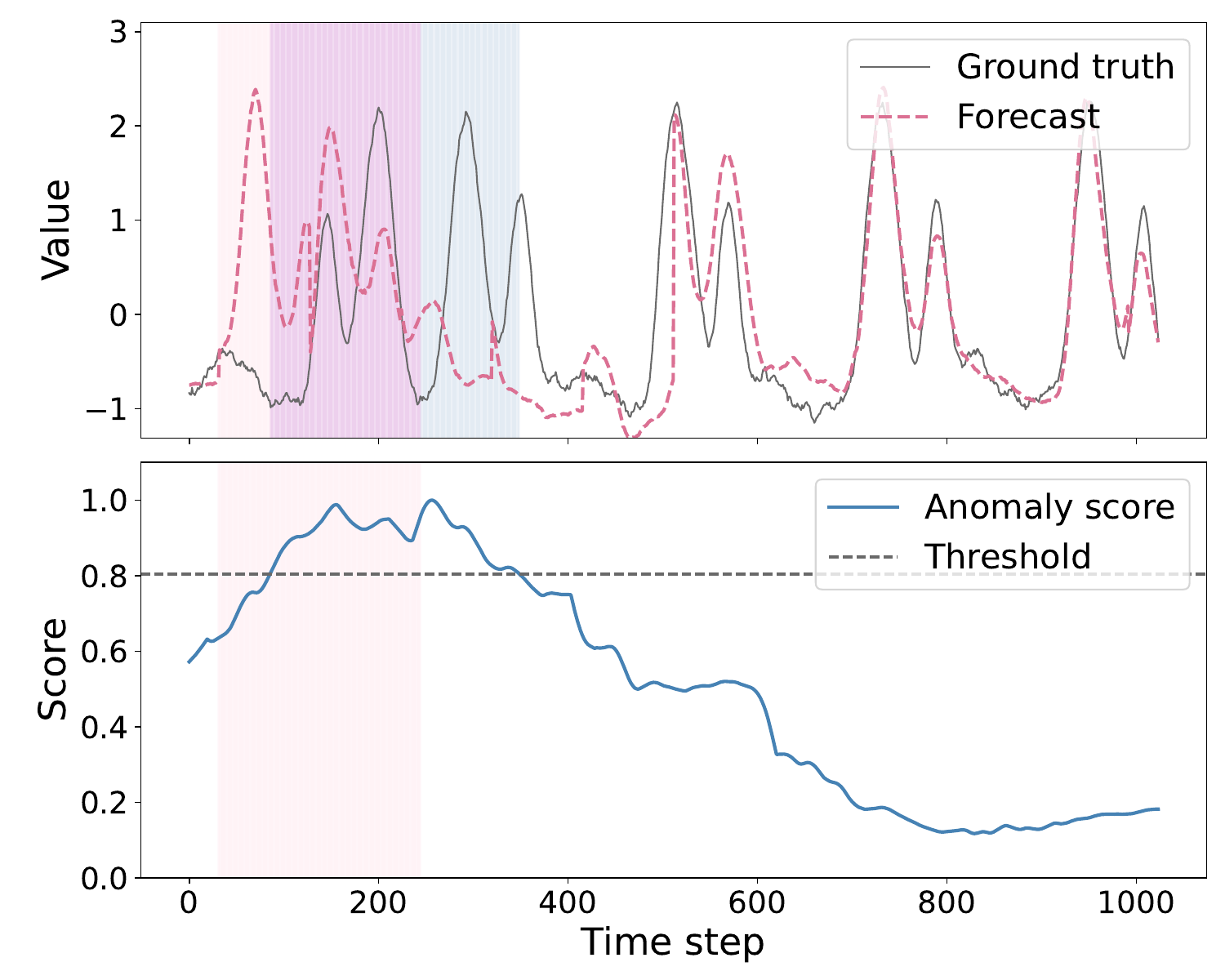}
        \caption{False positives caused by anomalies in the preceding time step.}
        \label{fig:subfig_c}
    \end{subfigure}
    \caption{Three error cases in anomaly detection using \textbf{Time-MoE} with \textbf{RATFM}.  
    (a) Target (bottom) and retrieved example (top). The blue-highlighted regions indicate false positive anomalies.  
    (b)(c) Time series forecasting results (top) and the corresponding anomaly scores (bottom). 
    The highlighted regions indicate anomalies as follows: purple for correctly detected anomalies, pink for missed anomalies, and blue for incorrectly classified as anomalies.}
    \label{fig:fail_example}
\end{figure}




To clarify the limitations of current methods, 
we show three common cases where \textbf{RATFM} failed anomaly detection (\figref{fail_example}).

\smallskip
\smallskip \noindent\textbf{Dissimilar future time series between target and example.}
Even when a retrieved example's input time series \maruthree{was} similar to the target, 
false positives occur\maruthree{red} if the \textbf{future} time series 
differ\maruthree{ed} between the target and example. As shown in \figref{subfig_a}, the input time series (solid lines) of the target (bottom) and the example (top) are similar. However, the generated 
forecast using the future time series of the example (dotted line in the top figure) shows a large deviation from the actual observations (light dotted line in the bottom figure), as illustrated by the dark red dotted line in the bottom figure.

\shoetsutwo{\smallskip \noindent\textbf{Small deviations from normal patterns.}
It was difficult to detect anomalies that exhibit high frequency but small deviations from normal patterns, 
as the anomaly score was influenced by the magnitude of deviation from the original data point.
\figref{subfig_b} shows the forecast results (top), the anomaly scores (bottom), and the detection threshold (dashed line). 
The model assigned insufficient anomaly scores to the pink-highlighted anomalous region.
}


\shoetsutwo{\smallskip \noindent\textbf{False positives caused by anomalies in the preceding time step.}}
When anomalous regions were included in the model input, they degraded the forecast accuracy \shoetsutwo{of the subsequent time series and led to high anomaly scores, resulting in false positives.}
\figref{subfig_c} shows a case where the model correctly detected the anomaly 
(purple) but the anomaly region affected the next forecast. As a result, the subsequent normal region was incorrectly flagged as anomalous 
(blue).

\section{Conclusion} \label{sec:conclusion}



\shoetsutwo{
In this study, we proposed 1) RATFM, a 
domain-independent scheme
for empowering time series foundation models to incorporate examples and 2) SMA-based post-processing for anomaly scores.
Experimental results demonstrated that Time-MoE and Moment with RATFM outperformed existing methods, including text-based LLMs, and achieved performance comparable to that obtained with in-domain settings. 
Visualization and analysis also supported the need for our simple SMA-based post-processing to resolve the peak problem in score-based anomaly detection.
}


As future work, we aim to explore the applicability of the example-based method beyond anomaly detection to a broader range of time series tasks.
\shoetsutwo{
Furthermore, enabling time series foundation models to understand diverse natural language instructions may lead to broader impact by facilitating their use across a wide range of domains and tasks. We will release all the code and results at \url{https://github.com/takamaruocha/RATFM}.
}

\newpage
\bibliographystyle{splncs04}
\bibliography{reference}




\newpage
\appendix

\section{Preliminaries: Anomaly Detection Flow}\label{sec:preliminaries}
In recent neural-based approaches, anomaly detection is typically conducted via time series forecasting or reconstruction tasks. The forecast-based approach trains a model to forecast future values accurately based on past input time series, whereas the reconstruction-based approach trains a model to reconstruct input time series with intentionally masked values.

\noindent \textbf{Forecast-based Approach} \quad
A pretrained model $f_{\text{forecast}}$ receives an input time series $X_{1:T} \in \mathbb{R}^T$ of $T$ historical time steps and forecasts the future $H$ steps, denoted by $\hat{X}_{T+1:T+H} \in \mathbb{R}^H$. Here, $T$ denotes the input length, and $H$ denotes the forecast horizon. 
\begin{equation} 
\label{eqn:process_predict} 
\begin{aligned} 
\hat{X}_{T+1:T+H} = f_{\text{forecast}}(X_{1:T}). 
\end{aligned} 
\end{equation}
For each data point in the forecast time series $\hat{X}_{T+1:T+H}$, the anomaly score $AS(x_{t})$ is computed as the absolute difference between the forecast $\hat{x}_{t}$ and the corresponding ground truth $x_{t}$:
\begin{equation} 
\label{eqn:AS_predict} 
\begin{aligned} 
AS(x_{t}) = |\hat{x}_t - x_t|. 
\end{aligned} 
\end{equation}

\noindent \textbf{Reconstruction-based Approach} \quad
A pretrained model $f_{\text{reconst}}$ reconstructs the input time series $X_{1:T+H}$, and the reconstructed time series is denoted as
\begin{equation}
\label{eqn:process_reconst}
\begin{aligned}
\hat{X}_{1:T+H} = f_{\text{reconst}}(X_{1:T+H}).
\end{aligned} 
\end{equation}
To ensure consistency with the forecast-based approach, we compute anomaly scores over the interval $X_{T+1:T+H}$.
For each data point in the reconstructed time series $\hat{X}_{T+1:T+H}$, the anomaly score $AS(x_{t})$ is computed as
\begin{equation}
\begin{aligned} 
AS(x_{t}) = |\hat{x}_t - x_t|.
\end{aligned}
\end{equation}


\section{Datasets} \label{appendix:datasets}
\begin{table}[t]
\centering
\caption{Details of the UCR Anomaly Archive.}
\label{tab:UCR_Anomaly_Archive}
\resizebox{\textwidth}{!}{
\begin{tabular}{l|r|rr|rr|rr}
\bhline{1.25pt}
\multicolumn{1}{c|}{\multirow{2}{*}{Domain}} & \multicolumn{1}{c|}{\multirow{2}{*}{\# Time Series}} & \multicolumn{2}{c|}{\# Data Points (Train)} & \multicolumn{2}{c|}{\# Data Points (Test)} & \multicolumn{2}{c}{Anomaly Length} \\
\multicolumn{1}{c|}{} & \multicolumn{1}{c|}{} & \multicolumn{1}{c}{Mean} & \multicolumn{1}{c|}{Median} & \multicolumn{1}{c}{Mean} & \multicolumn{1}{c|}{Median} & \multicolumn{1}{c}{Mean} & \multicolumn{1}{c}{Median} \\ \hline
ECG & 91 & 19726.19 & 15000.00 & 66454.00 & 35740.00 & 253.38 & 201.00 \\
Respiration & 17 & 51058.82 & 45000.00 & 144291.35 & 147000.00 & 169.94 & 2.00 \\
Gait & 33 & 35077.91 & 18500.00 & 84379.39 & 40500.00 & 317.91 & 214.00 \\
Satellite & 11 & 3500.00 & 3500.00 & 7845.09 & 7834.00 & 67.55 & 98.00 \\
Power Demand & 11 & 17922.64 & 16000.00 & 28329.91 & 15931.00 & 173.36 & 97.00 \\
Insect EPG & 25 & 4760.00 & 5000.00 & 16416.60 & 22826.00 & 73.88 & 51.00 \\
Arterial BP & 42 & 24684.43 & 3000.00 & 43844.31 & 6154.00 & 164.62 & 111.00 \\
Temperature & 13 & 4000.00 & 4000.00 & 4184.00 & 4184.00 & 34.38 & 34.38 \\
Accelerometer & 7 & 5485.71 & 2700.00 & 8905.29 & 5184.00 & 152.43 & 161.00 \\ \hline
All Domains & 250 & 21209.80 & 9406.00 & 56205.27 & 25001.00 & 197.45 & 101.00 \\ \bhline{1.25pt}
\end{tabular}
}
\end{table}

The details of the UCR Anomaly Archive are presented in \tabref{UCR_Anomaly_Archive}.
In many prior studies on anomaly detection, the reliability of evaluation results has been questioned owing to the inherent flaws in the datasets~\cite{wu2021current}.
The UCR Anomaly Archive was developed as a benchmark dataset to address issues such as imbalance in task difficulty, unrealistic anomaly density, mislabeling, and location bias in anomaly occurrences~\cite{10.14778/3685800.3685842}.
\section{Hyper-parameters} \label{appendix:hyper-parameters}
The hyper-parameter settings for \textbf{Time-MoE}, \textbf{Moment}, and \textbf{Anomaly Transformer} used in the evaluation experiments are summarized in \tabref{hyperparams}.
Except for the input length and forecast horizon, the hyper-parameter settings were based on those reported in the original papers.

\begin{table}[t]
\centering
\caption{Hyper-parameter settings for \textbf{Time-MoE}, \textbf{Moment}, and \textbf{Anomaly Transformer}.}
\label{tab:hyperparams}
\footnotesize
\begin{tabular}{l|l}
\toprule
\multicolumn{1}{c|}{Model} & \multicolumn{1}{c}{Hyper-parameters} \\
\midrule
\multirow{19}{*}{\textbf{Time-MoE}} 
& input length: 1120 \\
& forecast horizon: 96 \\
& learning rate: 1e-4 \\
& minimum learning rate: 5e-5 \\
& number of training epochs: 10 \\
& batch size: 64 \\
& warmup ratio: 0.0 \\
& warmup steps: 0 \\
& weight decay: 0.1 \\
& adam beta1: 0.9 \\
& adam beta2: 0.95 \\
& max gradient norm: 1.0 \\
& layers: 12 \\
& heads: 12 \\
& experts: 8 \\
& K (activated non-shared experts): 2 \\
& $d_\text{model}$: 768 \\
& $d_\text{ff}$: 3072 \\
& $d_\text{expert}$: 384 \\
\midrule
\multirow{17}{*}{\textbf{Moment}} 
& input length: 512 \\
& forecast horizon: 96 \\
& patch length: 8 \\
& patch stride length: 8 \\
& mask ratio: 30\% \\
& learning rate: 1e-4 \\
& minimum learning rate: 1e-5 \\
& number of training epochs: 10 \\
& batch size: 32 \\
& weight decay: 0.0 \\
& adam beta1: 0.9 \\
& adam beta2: 0.999 \\
& max gradient norm: 5.0 \\
& head dropout: 0.1 \\
& layers: 24 \\
& heads: 16 \\
& $d_\text{model}$: 1024 \\
& $d_\text{ff}$: 4096 \\
\midrule
\multirow{12}{*}{\textbf{Anomaly Transformer}} 
& input length: 512 \\
& k: 3 \\
& learning rate: 1e-4 \\
& number of training epochs: 10 \\
& batch size: 32 \\
& weight decay: 0.0 \\
& adam beta1: 0.9 \\
& adam beta2: 0.999 \\
& layers: 3 \\
& heads: 8 \\
& $d_\text{model}$: 512 \\
& $d_\text{ff}$: 512 \\
\bottomrule
\end{tabular}
\end{table}

\clearpage
\section{Prompts for GPT-4o}\label{appendix:GPTprompt}
\figref{gpt-4o} presents an example of a prompt used for anomaly detection with \textbf{GPT-4o}. In this prompt, \textbf{GPT-4o} was instructed to generate anomaly spans in a JSON format.
The generated output was then converted into binary labels, enabling direct comparison with the ground truth anomaly labels.
Following the experimental setting in prior work~\cite{zhou2025can}, we set the temperature to 0.4 and conducted the experiments in \textbf{Zero-shot} setting.

\begin{figure}[t]
\centering
\begin{tcolorbox}[enhanced,
  colback=gray!10,           
  colframe=gray!80!black,    
  title=\textbf{Prompt},
  coltitle=white,
  colbacktitle=gray!80!black,
  fonttitle=\bfseries,
  rounded corners,
  listing only,
  listing options={
    basicstyle=\ttfamily\small,
    breaklines=true,
    breakatwhitespace=true,
    showstringspaces=false
  }
]
-0.13 -0.17 -0.24 -0.25 -0.29 -0.26 -0.26 -0.26 -0.23 -0.17 -0.15 -0.16 -0.19 -0.22 -0.24 -0.29 -0.3 -0.33 -0.37 -0.33 -0.33 -0.32 -0.3 -0.29 -0.34 -0.37 -0.39 -0.45 -0.45 -0.43 -0.42 -0.45 -0.48 -0.48 -0.5 -0.43 -0.49 -0.44 -0.41 -0.41 -0.41 -0.44 -0.41 -0.42 -0.36 -0.31 -0.17 -0.07 0.0 0.05 0.14 0.19 0.24 0.17 -0.01 -0.17 -0.42 -0.77 -1.3 -2.02 -2.58 -3.05 -3.44 -3.68 -3.82 -3.83 -3.82 -3.69 -3.46 -3.19 -2.97 -2.71 -2.42 -2.17 -1.82 -1.47 -1.12 -0.92 -0.74 -0.6 -0.47 -0.34 -0.27 -0.17 -0.13 -0.08 -0.03 0.02 0.03 0.08 0.08 0.1 0.11 0.15 0.22 0.24
\\[1em]
Assume there are up to 5 anomalies. Detect ranges of anomalies in this time series, in terms of the x-axis coordinate.
List one by one, in JSON format. 
If there are no anomalies, answer with an empty list [].

Output template:
[\{"start": ..., "end": ...\}, \{"start": ..., "end": ...\}]
\end{tcolorbox}
\vspace{0.5em}
\captionof{figure}{Example of a prompt in \textbf{GPT-4o}.}
\label{fig:gpt-4o}
\end{figure}

\section{Forecasting vs. Reconstruction}\label{appendix:forecasting_vs_reconstruction}

\begin{figure*}[t]
    \centering
    \begin{subfigure}{0.33\textwidth} 
        \centering
        \includegraphics[width=\textwidth]{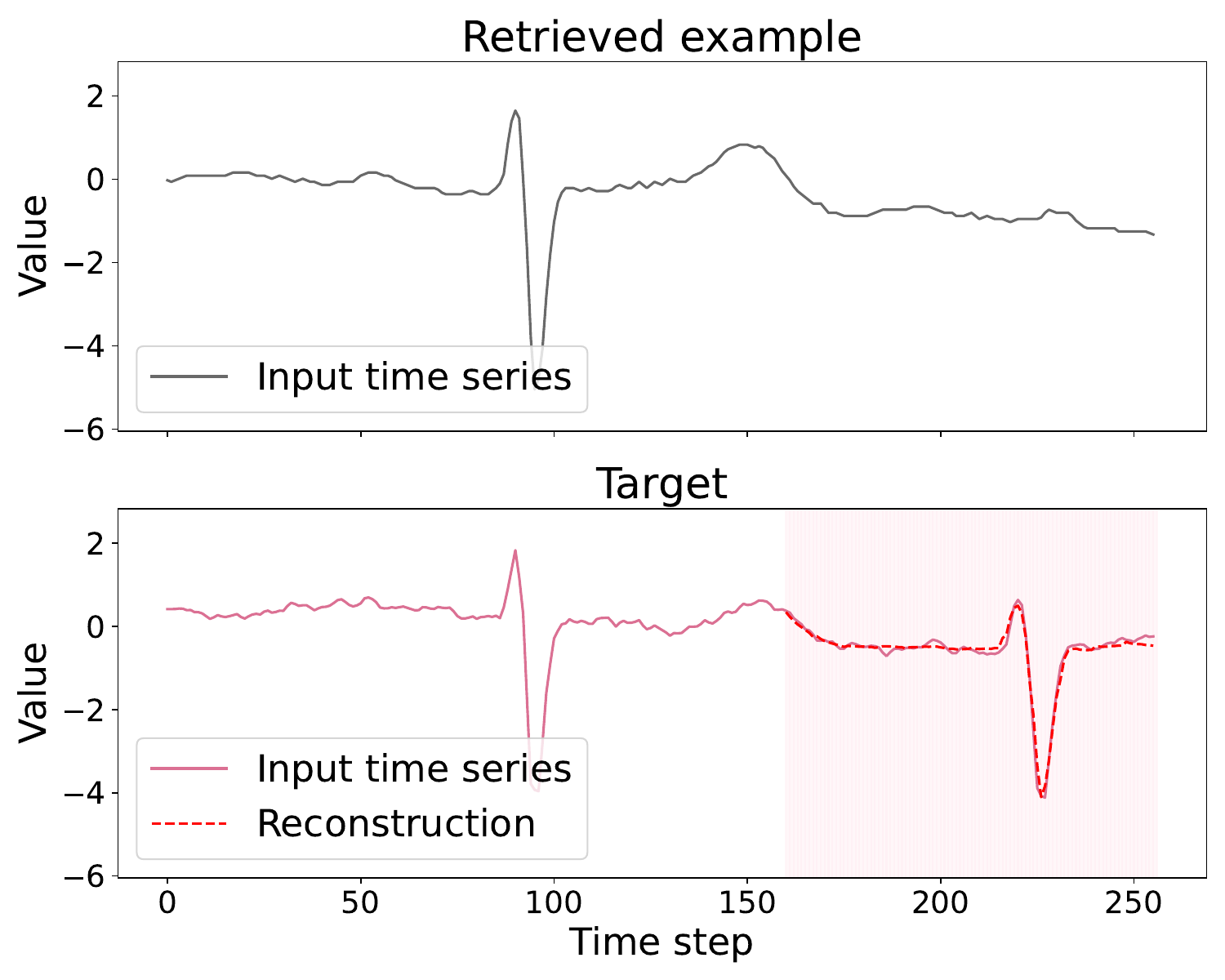}
    \end{subfigure}
    \hspace{-0.5em} 
    \begin{subfigure}{0.33\textwidth} 
        \centering
        \includegraphics[width=\textwidth]{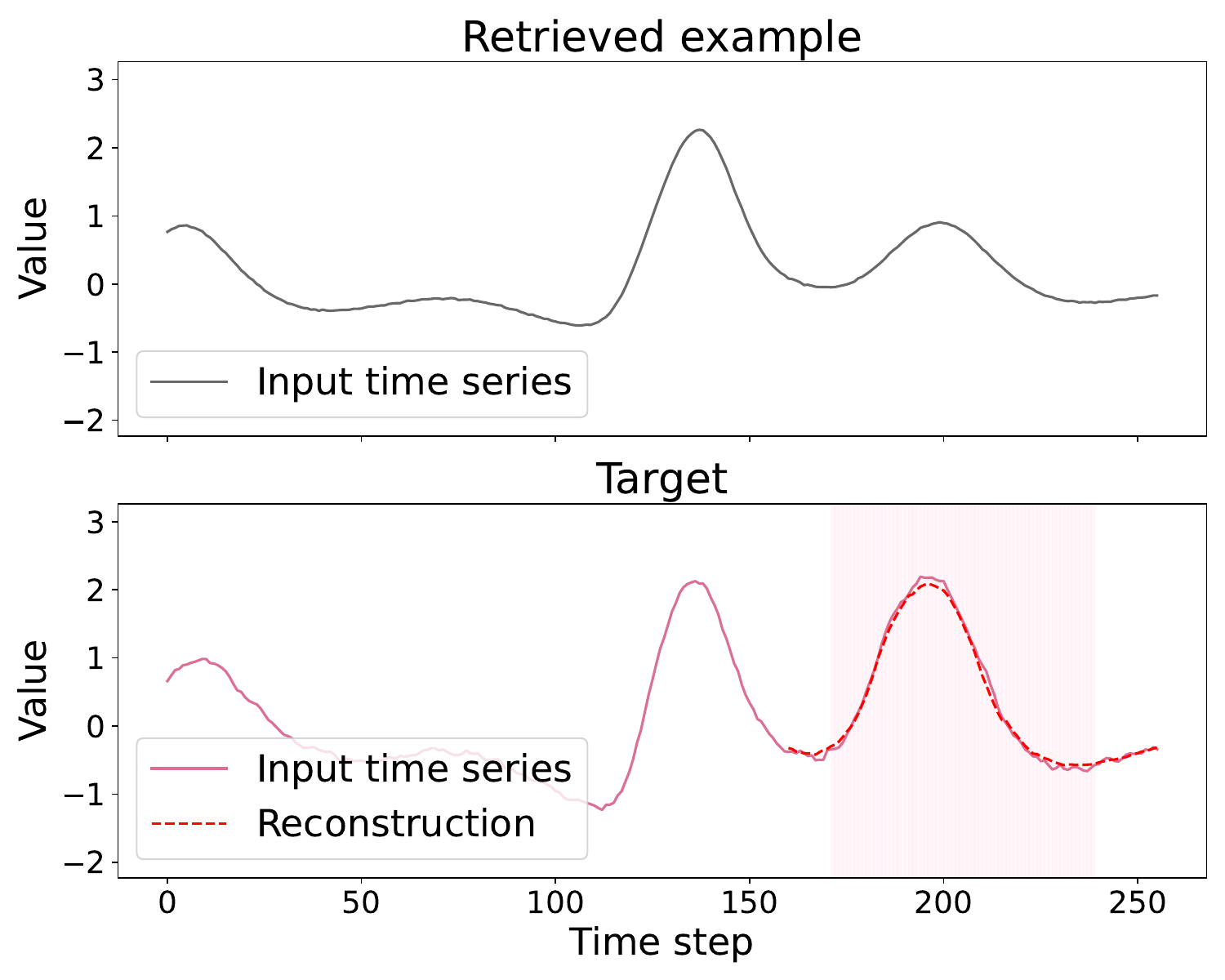}
    \end{subfigure}
    \hspace{-0.5em} 
    \begin{subfigure}{0.33\textwidth}
        \centering
        \includegraphics[width=\textwidth]{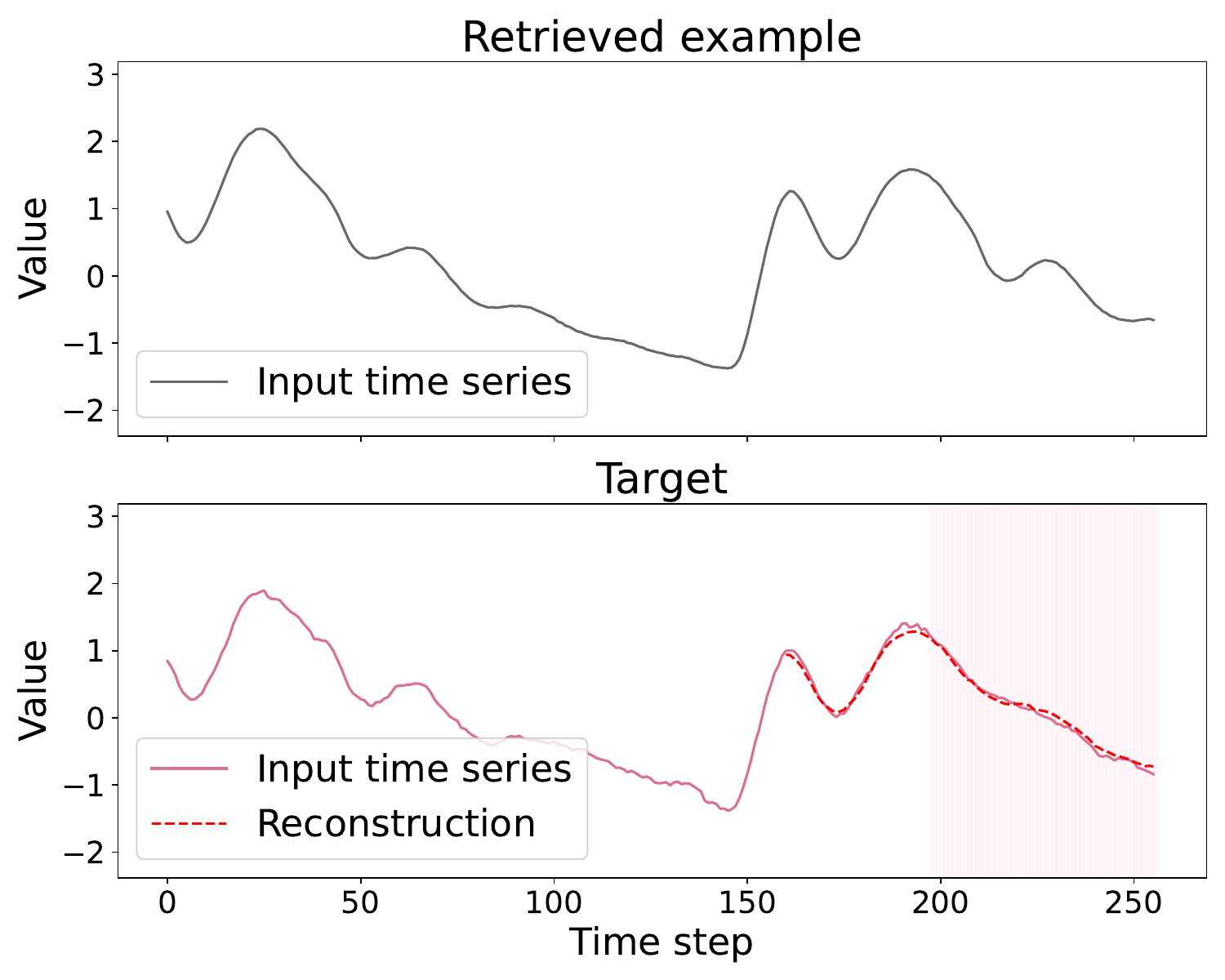}
    \end{subfigure}
    \caption{Target (bottom) and retrieved example (top) in \textbf{Moment} with \textbf{RATFM (reconstruction)}. The highlighted regions indicate ground-truth anomalies.}
    \label{fig:best_match_reconst}
\end{figure*}

As shown in \ssecref{main-results-vusroc}, \textbf{Moment} exhibited lower performance in \textbf{RATFM (reconstruction)} compared with \textbf{RATFM (forecast)} setting.
Unlike forecast-based methods, reconstruction-based approaches included the target interval $X_{T+1:T+H}$ as part of the input. As a result, the model reproduced this interval with high accuracy, even when it contained anomalies.
\figref{best_match_reconst} shows a target time series (bottom) and the retrieved example (top). Despite the example being included in the input, the model did not follow its pattern in the reconstruction, and the anomalous data points were fully reproduced. As a result, the reconstruction error was low, making the anomaly difficult to detect.
In contrast, forecast-based methods did not include $X_{T+1:T+H}$ in the input and were therefore not affected by this issue.

\begin{table}[t]
\footnotesize
\centering
\caption{Comparison of model performance without post-processing of anomaly scores.}
\label{tab:main-results-vusroc-noSMA}
\resizebox{\textwidth}{!}{
\begin{tabular}{llrrrrr}
\bhline{1.25pt}
\multicolumn{1}{c}{\multirow{2}{*}{Base Model}} & \multicolumn{1}{c}{\multirow{2}{*}{Training Setting}} & \multicolumn{1}{c}{\multirow{2}{*}{VUS-ROC}} & \multicolumn{1}{c}{\multirow{2}{*}{VUS-PR}} & \multicolumn{3}{c}{Point-wise} \\
\multicolumn{1}{c}{} & \multicolumn{1}{c}{} & \multicolumn{1}{c}{} & \multicolumn{1}{c}{} & \multicolumn{1}{c}{F1 Score} & \multicolumn{1}{c}{Precision} & \multicolumn{1}{c}{Recall} \\ \hline
\multirow{5}{*}{\textbf{Time-MoE}} 
 & \textbf{Zero-shot} & 65.7\% & 5.5\% & 5.7\% & 6.3\% & 10.4\% \\  
 & \textbf{Out-domain FT} & 65.0\% & 4.9\% & 5.2\% & 5.8\% & 10.1\% \\  
 & \textbf{In-domain FT} & 70.3\% & 7.9\% & 8.8\% & 9.0\% & 15.3\% \\  
 & \textbf{RATFM} & 68.9\% & 6.8\% & 7.4\% & 7.5\% & 13.2\% \\  
 & \textbf{RATFM w/o training} & 65.0\% & 5.7\% & 6.0\% & 6.6\% & 11.1\% \\ \hline  
\multirow{6}{*}{\textbf{Moment}} 
 & \textbf{Zero-shot} & 64.9\% & 3.3\% & 3.4\% & 3.3\% & 9.3\% \\  
 & \textbf{Out-domain FT} & 65.1\% & 3.3\% & 3.1\% & 3.0\% & 9.0\% \\  
 & \textbf{In-domain FT} & 67.5\% & 5.6\% & 5.8\% & 5.6\% & 13.7\% \\  
 & \textbf{RATFM (forecast)} & 68.8\% & 6.7\% & 6.9\% & 7.4\% & 11.7\% \\  
 & \textbf{RATFM w/o training} & 64.2\% & 3.4\% & 3.5\% & 3.6\% & 9.0\% \\  
 & \textbf{RATFM (reconstruction)} & 63.7\% & 5.0\% & 4.3\% & 4.5\% & 10.6\% \\ \hline  
\multirow{2}{*}{\textbf{Anomaly Transformer}} 
 & \textbf{Time Series-wise} & 51.3\% & 1.4\% & 0.2\% & 0.1\% & 0.2\% \\  
 & \textbf{In-domain} & 50.3\% & 2.6\% & 0.3\% & 1.6\% & 0.4\% \\ \hline  
\multirow{2}{*}{\textbf{Sub-PCA}} 
 & \textbf{Time Series-wise} & 63.3\% & 3.1\% & 2.8\% & 3.1\% & 6.9\% \\  
 & \textbf{In-domain} & 61.1\% & 2.6\% & 2.1\% & 3.2\% & 5.5\% \\ \hline  
\textbf{GPT-4o} & \textbf{Zero-shot} & -- & -- & 1.5\% & 1.0\% & 8.9\% \\ \bhline{1.25pt}
\end{tabular}
}
\end{table}

\begin{table}[t]
\centering
\scriptsize
\caption{Comparison of model performance across individual domains based on VUS-ROC after applying SMA.}
\label{tab:main-result-domain}
\begin{tabular}{
  L{2cm}
  L{3cm}
  R{1.2cm} R{1.2cm} R{1.2cm} R{1.2cm} R{1.2cm}
}
\bhline{1.25pt}
\toprule
\makecell{Model} & \makecell{Training Setting} & \makecell{ECG} & \makecell{Resp.} & \makecell{Gait} & \makecell{Satellite} & \makecell{Power\\Demand} \\
\midrule
\multirow{5}{*}{\textbf{Time-MoE}} 
& \textbf{Zero-shot}              & 67.7\% & 58.4\% & 60.8\% & 63.8\% & 69.6\% \\
& \textbf{Out-domain FT}          & 70.4\% & 73.2\% & 65.8\% & 89.1\% & 74.4\% \\
& \textbf{In-domain FT}           & 81.3\% & 62.6\% & 80.9\% & 92.7\% & 78.4\% \\
& \textbf{RATFM}                  & 73.9\% & 75.8\% & 74.1\% & 93.0\% & 68.4\% \\
& \textbf{RATFM w/o training}     & 65.3\% & 42.6\% & 55.9\% & 74.2\% & 70.4\% \\
\midrule
\multirow{6}{*}{\textbf{Moment}}
& \textbf{Zero-shot}              & 74.4\% & 52.6\% & 64.1\% & 95.1\% & 64.0\% \\
& \textbf{Out-domain FT}          & 75.1\% & 51.4\% & 67.4\% & 90.7\% & 62.3\% \\
& \textbf{In-domain FT}           & 77.6\% & 54.7\% & 76.0\% & 91.5\% & 71.8\% \\
& \textbf{RATFM (forecast)}                  & 73.3\% & 67.5\% & 75.8\% & 82.8\% & 64.1\% \\
& \textbf{RATFM w/o training}     & 72.8\% & 35.2\% & 64.1\% & 88.1\% & 60.5\% \\
& \textbf{RATFM (reconstruction)} & 74.4\% & 52.6\% & 64.1\% & 95.1\% & 64.0\% \\
\midrule
\multirow{2}{*}{\makecell[l]{\textbf{Anomaly}\\\textbf{Transformer}}}
& \textbf{Time Series-wise}       & 52.0\% & 41.3\% & 49.9\% & 43.6\% & 59.1\% \\
& \textbf{In-domain}              & 52.9\% & 50.1\% & 51.7\% & 51.1\% & 59.1\% \\
\midrule
\multirow{2}{*}{\textbf{Sub-PCA}}
& \textbf{Time Series-wise}       & 62.3\% & 63.1\% & 58.6\% & 77.3\% & 55.4\% \\
& \textbf{In-domain}              & 60.5\% & 60.9\% & 49.5\% & 76.9\% & 58.0\% \\
\bottomrule
\bhline{1.25pt}
\end{tabular}

\vspace{1em}

\begin{tabular}{
  L{2cm}
  L{3cm}
  R{1.2cm} R{1.2cm} R{1.2cm} R{1.2cm} R{1.2cm}
}
\bhline{1.25pt}
\toprule
\makecell{Model} & \makecell{Training Setting} & \makecell{Insect\\EPG} & \makecell{Arterial\\BP} & \makecell{Temp.} & \makecell{Accel.} & \makecell{Average} \\
\midrule
\multirow{5}{*}{\textbf{Time-MoE}} 
& \textbf{Zero-shot}              & 68.7\% & 74.7\% & 84.0\% & 73.8\% & 68.7\% \\
& \textbf{Out-domain FT}          & 67.7\% & 67.3\% & 79.0\% & 64.7\% & 70.5\% \\
& \textbf{In-domain FT}           & 72.1\% & 82.6\% & 77.7\% & 70.7\% & 79.1\% \\
& \textbf{RATFM (forecast)}                  & 72.3\% & 79.2\% & 84.9\% & 78.1\% & 76.1\% \\
& \textbf{RATFM w/o training}     & 66.8\% & 73.7\% & 83.9\% & 75.9\% & 65.9\% \\
\midrule
\multirow{6}{*}{\textbf{Moment}}
& \textbf{Zero-shot}              & 74.4\% & 63.4\% & 87.1\% & 73.4\% & 70.6\% \\
& \textbf{Out-domain FT}          & 72.5\% & 62.9\% & 87.4\% & 73.8\% & 70.7\% \\
& \textbf{In-domain FT}           & 79.0\% & 80.7\% & 94.7\% & 71.1\% & 77.4\% \\
& \textbf{RATFM (forecast)}                  & 74.4\% & 76.2\% & 79.2\% & 79.6\% & 74.3\% \\
& \textbf{RATFM w/o training}     & 72.6\% & 67.4\% & 87.6\% & 76.6\% & 69.0\% \\
& \textbf{RATFM (reconstruction)} & 74.4\% & 63.4\% & 87.1\% & 73.4\% & 67.1\% \\
\midrule
\multirow{2}{*}{\makecell[l]{\textbf{Anomaly}\\\textbf{Transformer}}}
& \textbf{Time Series-wise}       & 49.4\% & 55.6\% & 54.4\% & 66.3\% & 51.8\% \\
& \textbf{In-domain}              & 54.2\% & 55.4\% & 53.5\% & 57.6\% & 53.5\% \\
\midrule
\multirow{2}{*}{\textbf{Sub-PCA}}
& \textbf{Time Series-wise}       & 76.0\% & 62.8\% & 73.9\% & 94.2\% & 64.9\% \\
& \textbf{In-domain}              & 70.7\% & 57.9\% & 74.9\% & 68.1\% & 61.3\% \\ 
\bottomrule
\bhline{1.25pt}
\end{tabular}
\end{table}

\begin{table}[t]
\centering
\footnotesize
\caption{Similarity between the forecast target and the three types of input (\textbf{Time-MoE}). (a) the example future time series in \textbf{RATFM} 
(b) the segment of \textbf{Zero-shot} that corresponds to the example future time series in \textbf{RATFM} 
(c) the segment of \textbf{Zero-shot} that corresponds to the entire example time series in \textbf{RATFM}.}
\label{tab:similarity_domain}
\begin{tabular}{lrrr}
\bhline{1.25pt}
\multirow{2}{*}{Domain} & \multicolumn{1}{c}{\textbf{RATFM}} & \multicolumn{2}{c}{\textbf{Zero-shot}} \\
 & \multicolumn{1}{c}{(a)} & \multicolumn{1}{c}{(b)} & \multicolumn{1}{c}{(c)} \\ \hline
ECG            & $0.882$ & $0.013$   & $0.477$ \\
Respiration    & $0.989$ & $-0.027$  & $0.944$ \\
Gait           & $0.951$ & $0.002$   & $0.733$ \\
Satellite      & $0.994$ & $-0.004$  & $0.995$ \\
Power Demand   & $0.996$ & $-0.011$  & $0.899$ \\
Insect EPG     & $0.995$ & $-0.013$  & $0.579$ \\
Atrial BP      & $0.996$ & $-0.020$  & $0.835$ \\
Temperature    & $0.995$ & $-0.038$  & $0.788$ \\
Accelerometer  & $0.995$ & $-0.034$  & $0.660$ \\ \hline
Average        & $0.974$ & $-0.015$  & $0.768$ \\ 
\bhline{1.25pt}
\end{tabular}
\end{table}
\begin{table}[t]
\footnotesize
\centering
\caption{Comparison of model performance after applying SMA, evaluated using bootstrapping~\cite{eforn1979bootstrap}. The mean and standard deviation were estimated from 1,000 bootstrap samples of the evaluation scores across all time series.}
\label{tab:main-results-vusroc-error-bar}
\resizebox{\textwidth}{!}{
\begin{tabular}{llrrr}
\bhline{1.25pt}
\multicolumn{1}{c}{\multirow{2}{*}{Base Model}} & \multicolumn{1}{c}{\multirow{2}{*}{Training Setting}} & \multicolumn{1}{c}{\multirow{2}{*}{VUS-ROC}} & 
\multicolumn{1}{c}{\multirow{2}{*}{VUS-PR}} & 
\multicolumn{1}{c}{\multirow{2}{*}{F1 Score}} \\
\multicolumn{1}{c}{} & \multicolumn{1}{c}{} & \multicolumn{1}{c}{} & \multicolumn{1}{c}{} & \multicolumn{1}{c}{} \\ \hline
\multirow{5}{*}{\textbf{Time-MoE}} 
 & \textbf{Zero-shot} & $68.3\% \pm 1.8\%$ & $15.3\% \pm 1.7\%$& $11.4\% \pm 1.6\%$ \\
 & \textbf{Out-domain FT} & $70.3\% \pm 1.7\% $ & $13.9\% \pm 1.6\% $ & $9.7\% \pm 1.3\% $ \\
 & \textbf{In-domain FT} & $79.0\% \pm 1.3\% $ & $20.5\% \pm 1.9\% $ & $16.1\% \pm 1.6\% $ \\
 & \textbf{RATFM} & $76.2\% \pm 1.4\% $ & $17.7\% \pm 1.8\% $ & $13.2\% \pm 1.6\% $ \\ 
 & \textbf{RATFM w/o training} & $65.9\% \pm 1.8\% $ & $15.7\% \pm 1.8\% $ & $11.8\% \pm 1.5\% $ \\ \hline
\multirow{6}{*}{\textbf{Moment}} 
 & \textbf{Zero-shot} & $70.6\% \pm 1.7\%$ & $15.1\% \pm 1.5\%$ & $9.9\% \pm 1.2\%$ \\
 & \textbf{Out-domain FT} & $70.7\% \pm 1.7\% $ & $16.1\% \pm 1.5\% $ & $11.4\% \pm 1.3\% $ \\
 & \textbf{In-domain FT} & $77.4\% \pm 1.7\% $ & $21.7\% \pm 1.7\% $ & $14.3\% \pm 1.4\% $ \\
 & \textbf{RATFM (forecast)} & $74.3\% \pm 1.5\% $ & $16.3\% \pm 1.7\% $ & $12.6\% \pm 1.6\% $ \\
 & \textbf{RATFM w/o training} & $69.0\% \pm 1.8\%$ & $16.1\% \pm 1.5\%$ & $9.7\% \pm 1.3\%$ \\
 & \textbf{RATFM (reconstruction)} & $67.1\% \pm 1.8\% $ & $15.8\% \pm 1.6\% $ & $10.0\% \pm 1.3\% $ \\ \hline
\multirow{2}{*}{\textbf{Anomaly Transformer}} 
 & \textbf{Time Series-wise} & $51.8\% \pm 1.5\% $ & $2.9\% \pm 0.5\% $ & $1.2\% \pm 0.5\% $ \\
 & \textbf{In-domain} & $53.5\% \pm 1.3\% $ & $4.0\% \pm 0.7\% $ & $2.4\% \pm 0.6\% $ \\ \hline
\multirow{2}{*}{\textbf{Sub-PCA}} 
 & \textbf{Time Series-wise} & $64.9\% \pm 1.6\% $ & $11.3\% \pm 1.4\% $ & $7.8\% \pm 1.3\% $ \\
 & \textbf{In-domain} & $61.3\% \pm 1.6\% $ & $9.4\% \pm 1.3\% $ & $6.7\% \pm 1.2\% $ \\ \hline
\textbf{GPT-4o} & \textbf{Zero-shot} & -- & -- & $1.5\% \pm 0.1\% $ \\ \bhline{1.25pt}
\end{tabular}
}
\end{table}

\clearpage
\section{Other test cases}\label{appendix:other_test_cases}

\begin{figure*}[t]
    \centering
    \begin{subfigure}{\textwidth}
        \centering
        \begin{subfigure}{0.33\textwidth}
            \centering
            \includegraphics[width=\textwidth]{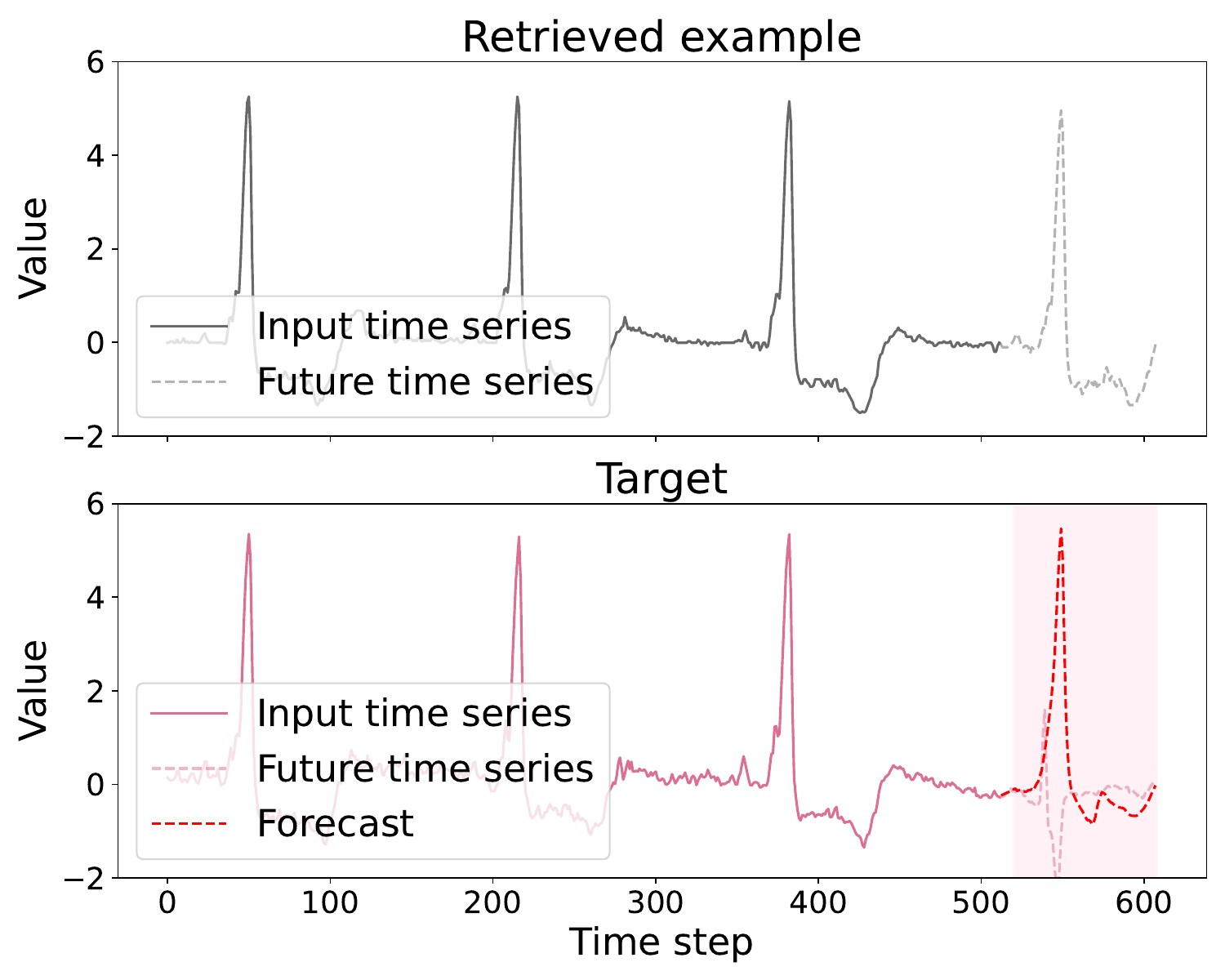}
        \end{subfigure}
        \hspace{-0.5em}
        \begin{subfigure}{0.33\textwidth}
            \centering
            \includegraphics[width=\textwidth]{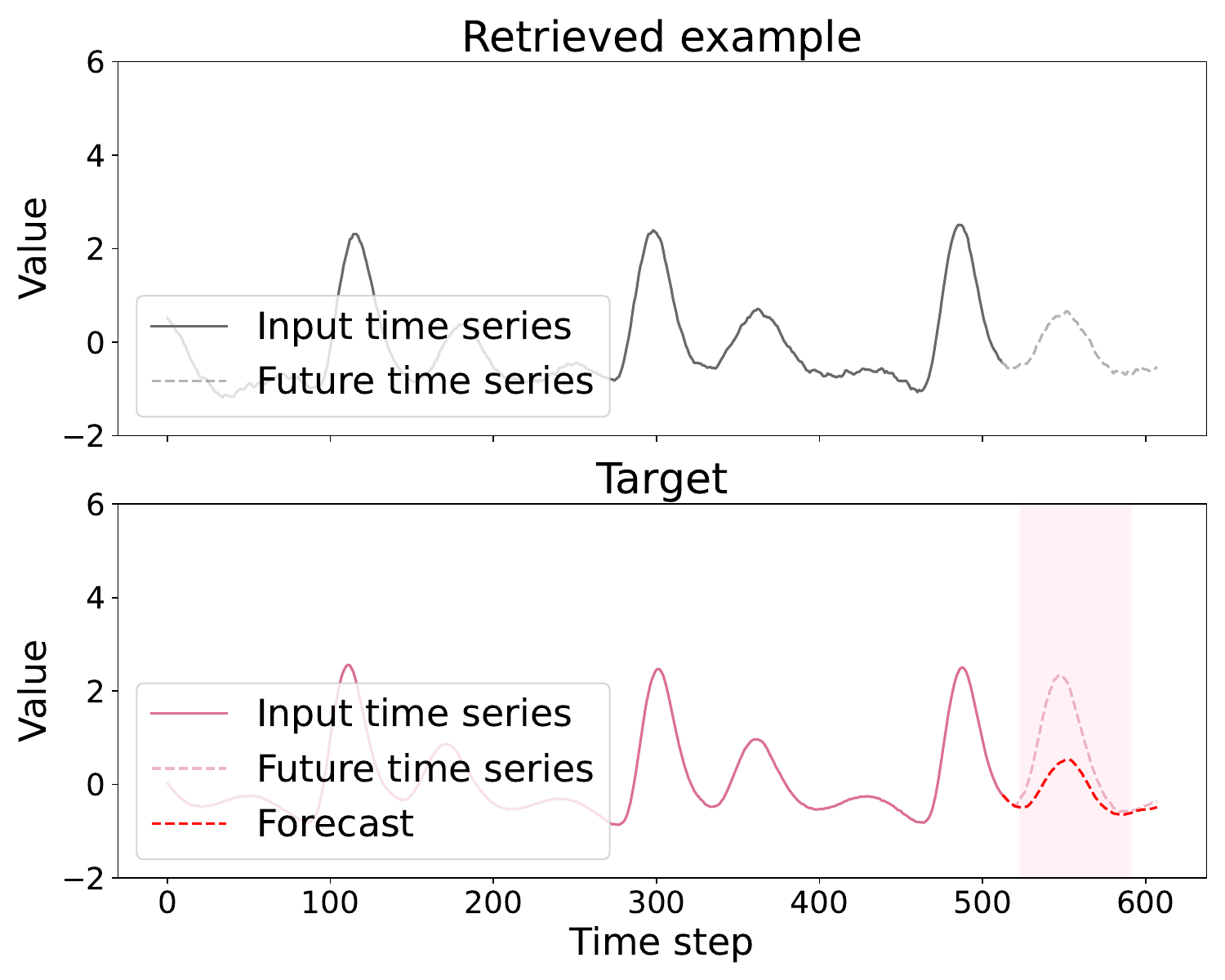}
        \end{subfigure}
        \hspace{-0.5em}
        \begin{subfigure}{0.33\textwidth}
            \centering
            \includegraphics[width=\textwidth]{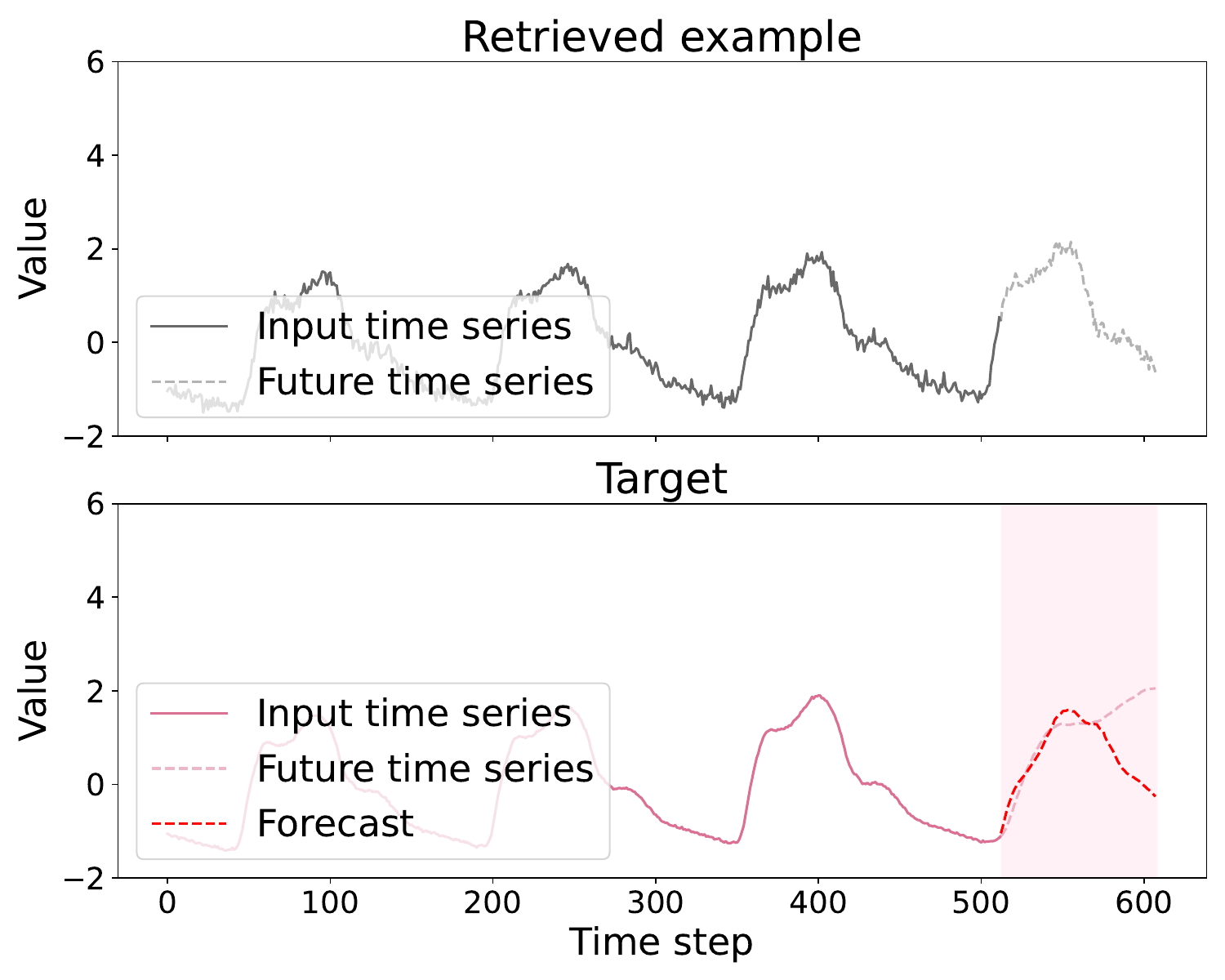}
        \end{subfigure}
        \caption{\textbf{Time-MoE}.}
    \end{subfigure}
    
    \vspace{1em} 

    \begin{subfigure}{\textwidth}
        \centering
        \begin{subfigure}{0.33\textwidth}
            \centering
            \includegraphics[width=\textwidth]{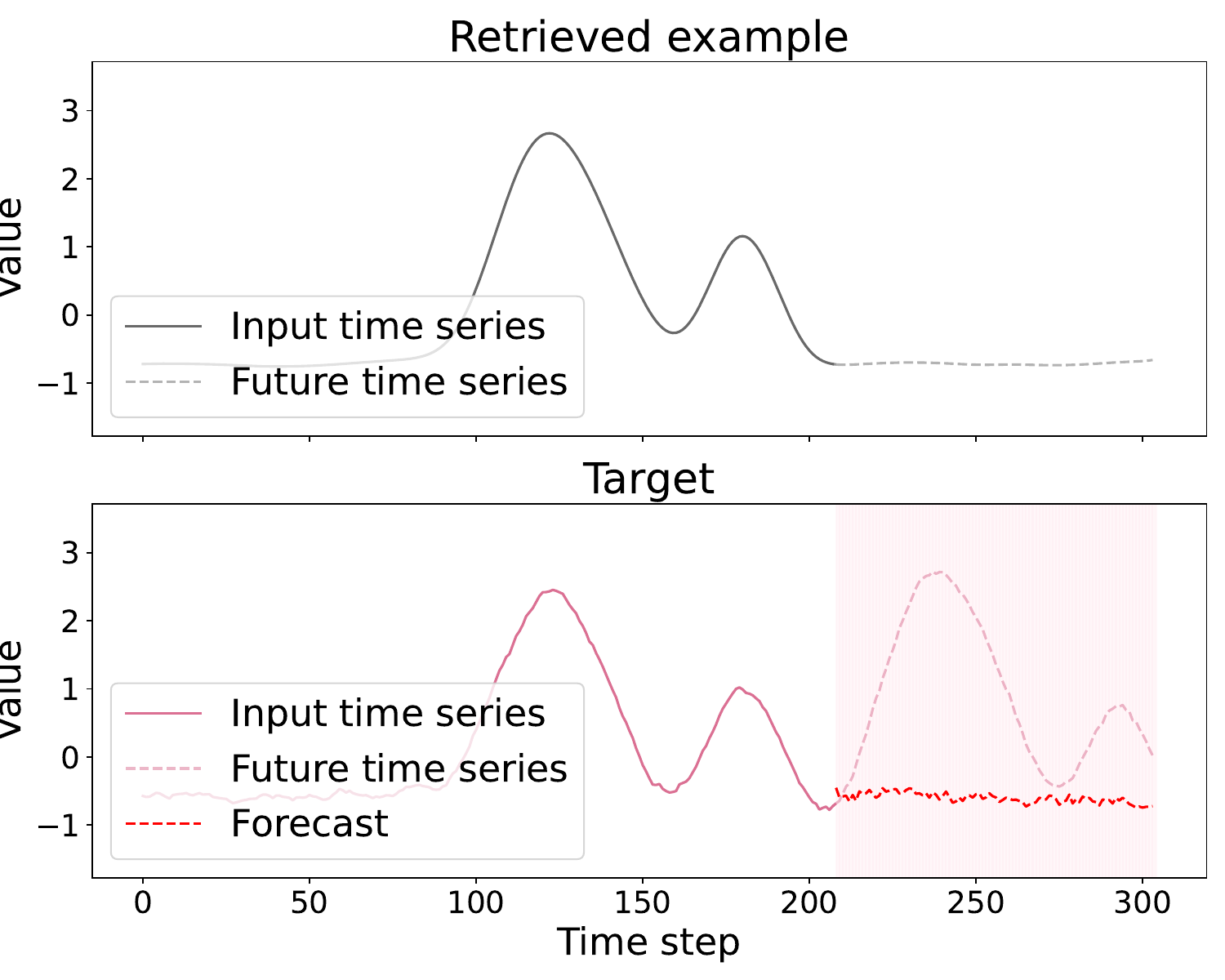}
        \end{subfigure}
        \hspace{-0.5em}
        \begin{subfigure}{0.33\textwidth}
            \centering
            \includegraphics[width=\textwidth]{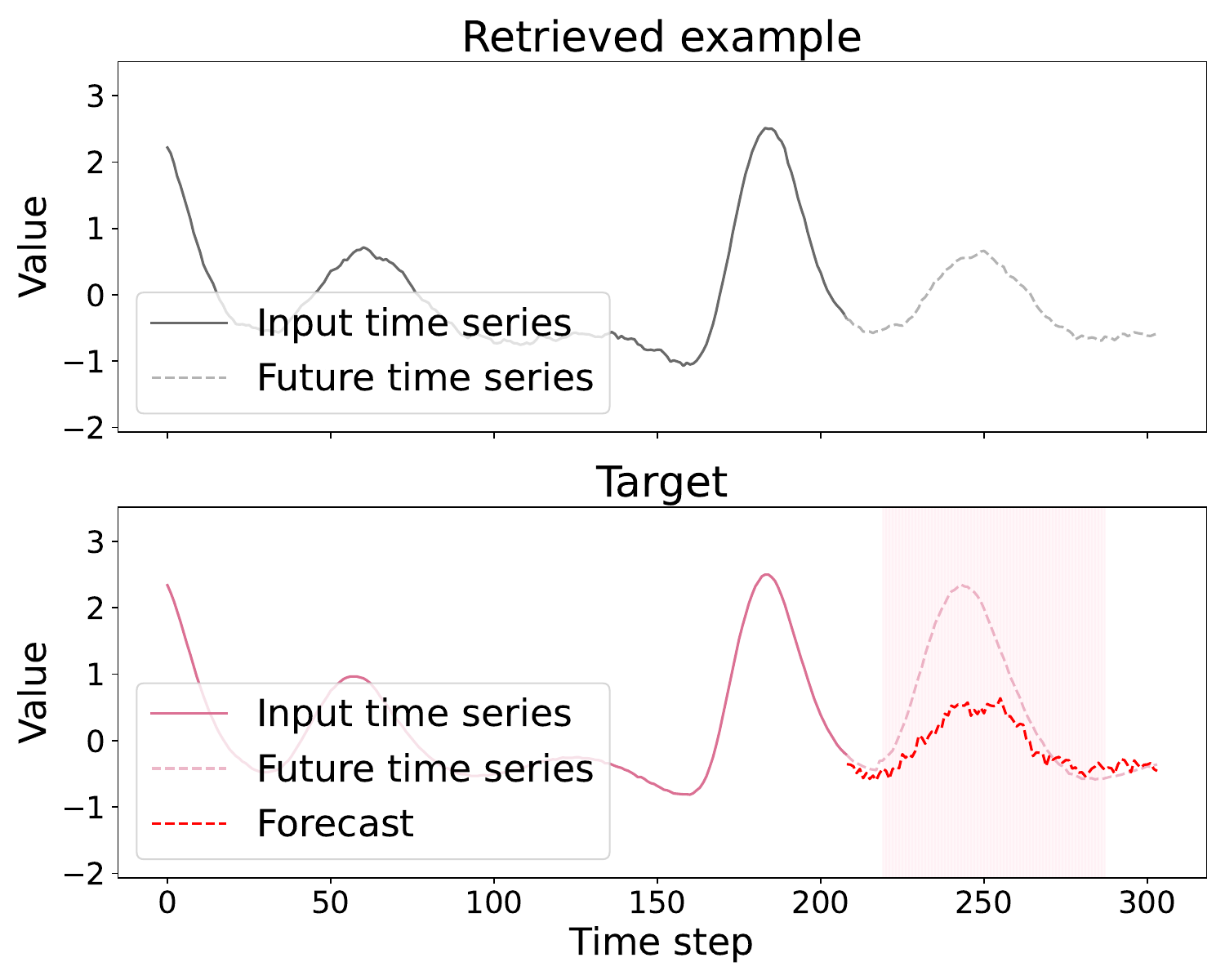}
        \end{subfigure}
        \hspace{-0.5em}
        \begin{subfigure}{0.33\textwidth}
            \centering
            \includegraphics[width=\textwidth]{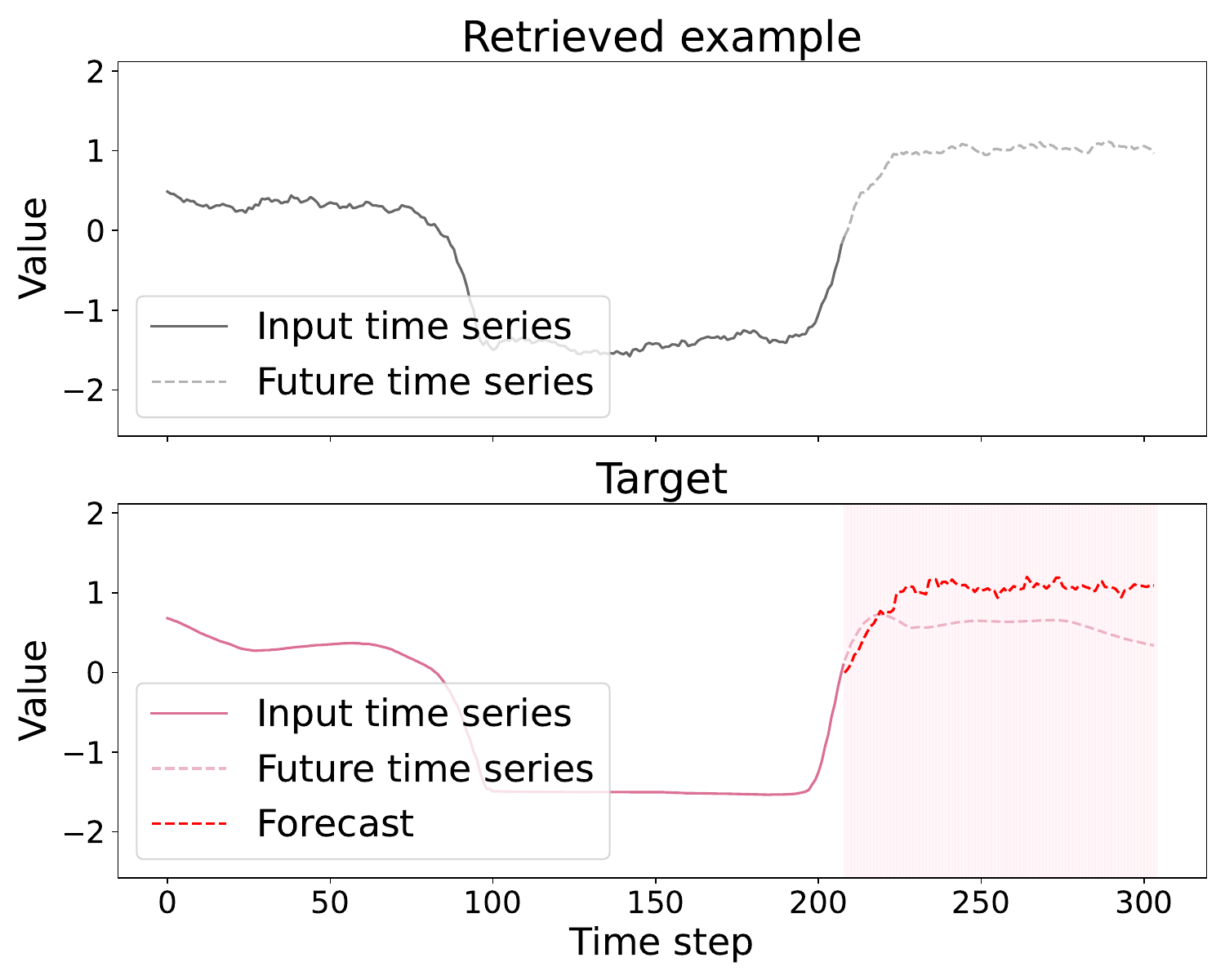}
        \end{subfigure}
        \caption{\textbf{Moment}.}
    \end{subfigure}
    
    \caption{Target (bottom) and retrieved example (top) under the \textbf{RATFM} setting. The highlighted regions indicate ground-truth anomalies.}
    \label{fig:best_match_examples_combined}
\end{figure*}

\figref{best_match_examples_combined} shows the target and retrieved examples under \textbf{Time-MoE} and \textbf{Moment} with \textbf{RATFM}.
Each figure illustrates a target (bottom) and a retrieved example (top). The solid lines represent the input time series, faint dashed lines indicate the future time series. In the bottom part, the bold red dashed lines show the forecasts generated by each model with \textbf{RATFM}.  
The highlighted regions denote the ground-truth anomalies. 
The fine-tuned models utilized the retrieved examples to perform time series forecasting and accurately detected anomalies based on the deviations from actual observations.

\begin{figure*}[t]
    \centering

    \begin{subfigure}{\textwidth}
        \centering
        \begin{subfigure}{0.33\textwidth}
            \centering
            \includegraphics[width=\textwidth]{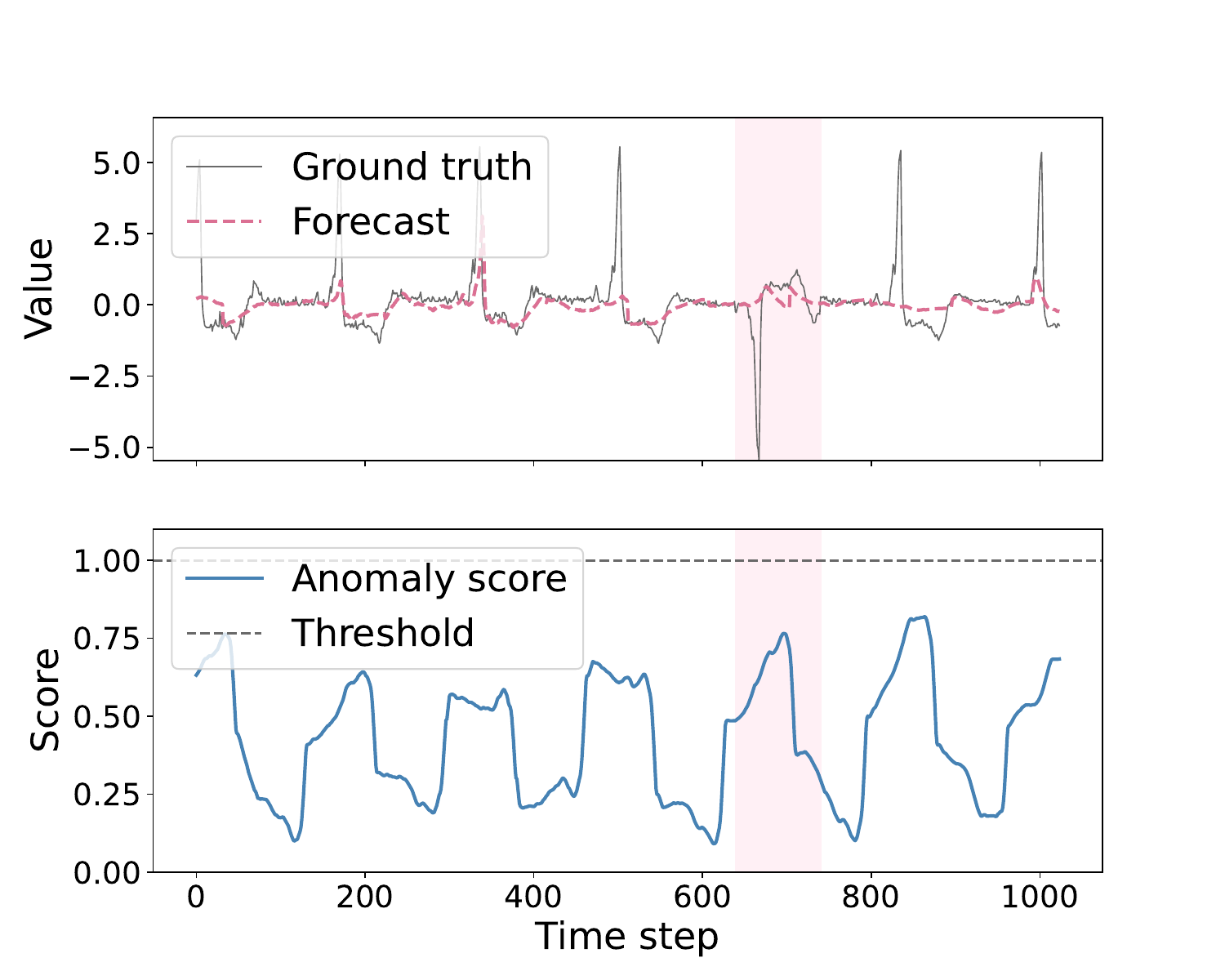}
            \caption*{\textbf{Zero-shot}.}
        \end{subfigure}
        \hspace{-0.01\textwidth}
        \begin{subfigure}{0.33\textwidth}
            \centering
            \includegraphics[width=\textwidth]{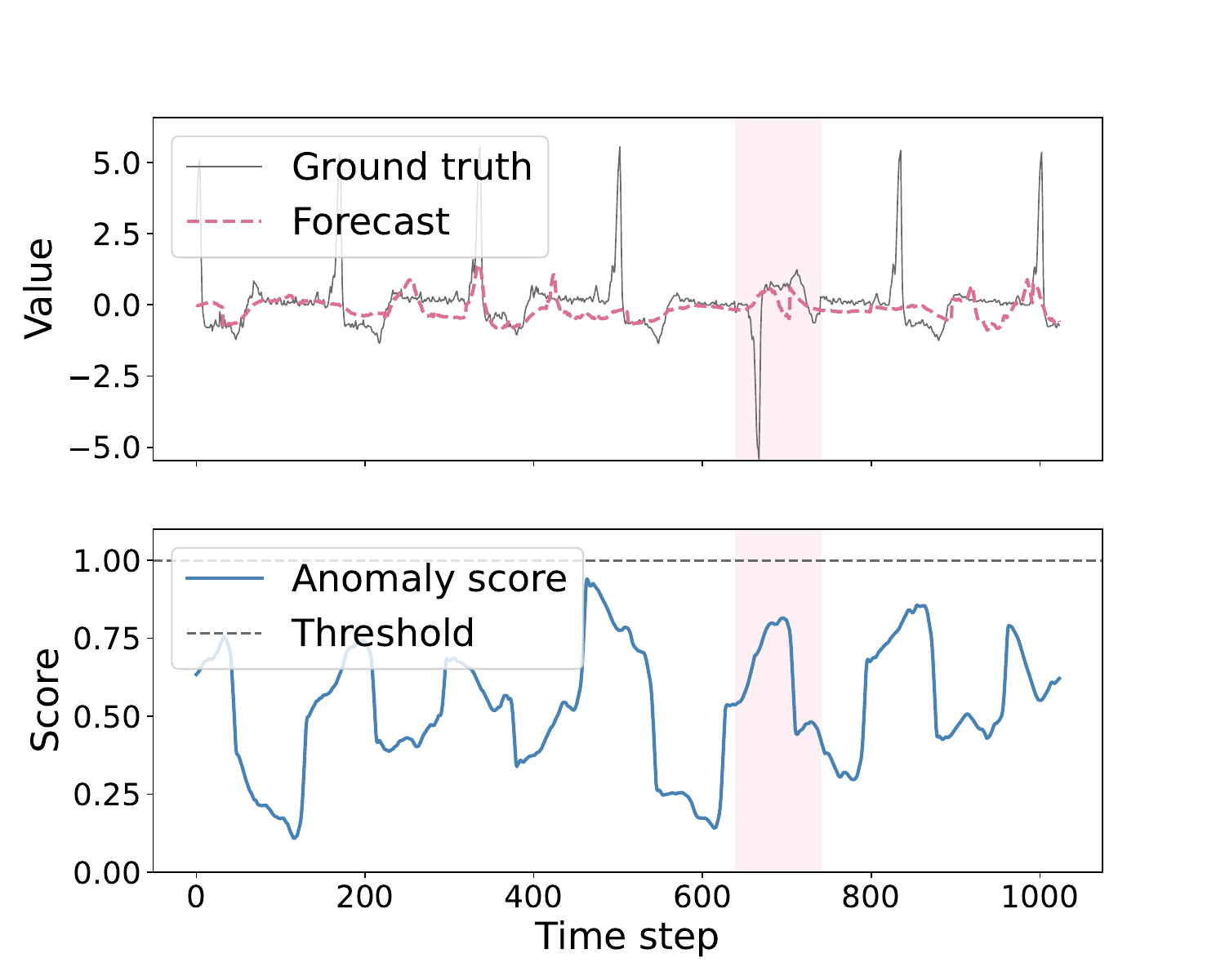}
            \caption*{\textbf{Out-domain FT}.}
        \end{subfigure}
        \hspace{-0.01\textwidth}
        \begin{subfigure}{0.33\textwidth}
            \centering
            \includegraphics[width=\textwidth]{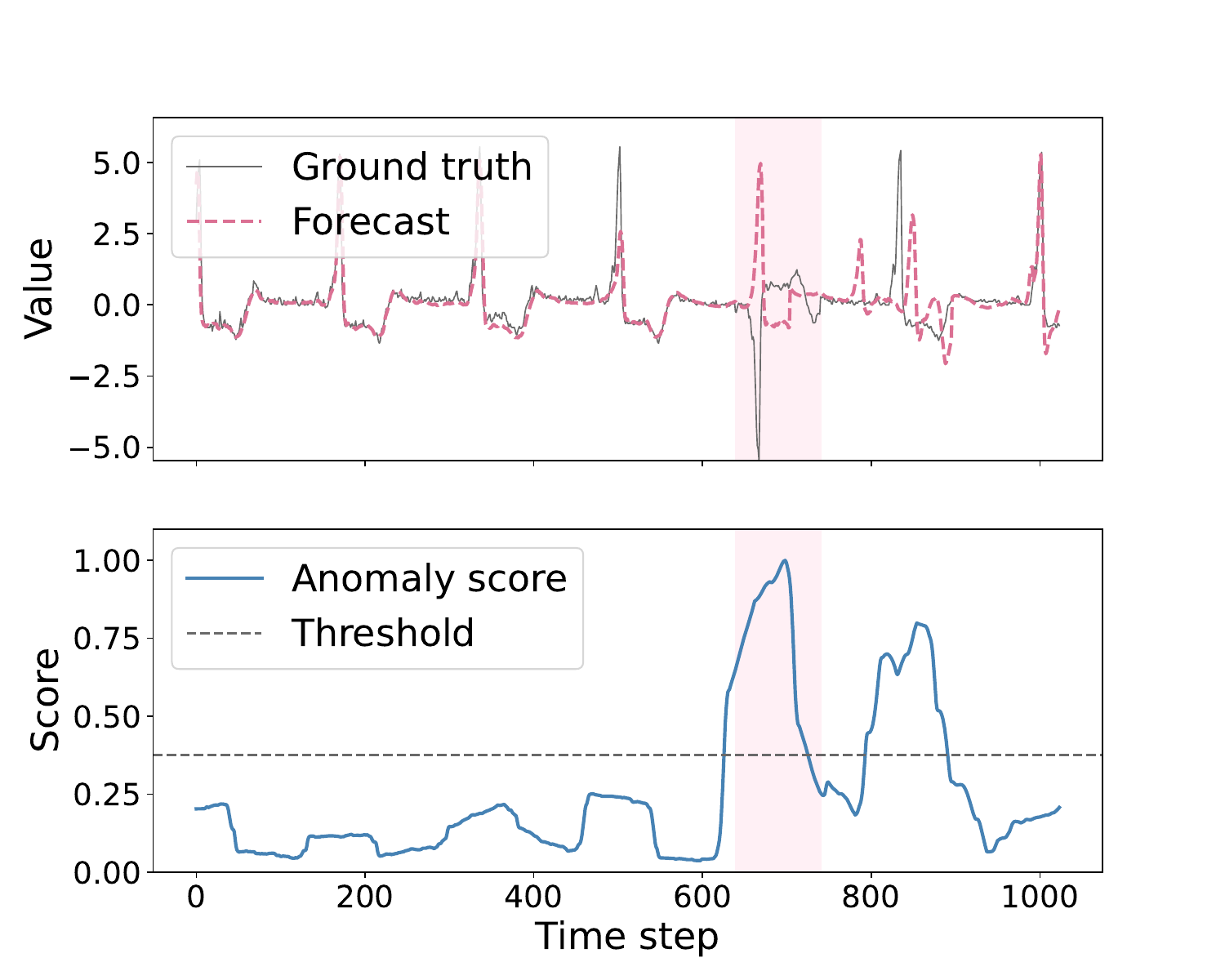}
            \caption*{\textbf{In-domain FT}.}
        \end{subfigure}

        \vspace{0.3em}

        \begin{subfigure}{0.33\textwidth}
            \centering
            \includegraphics[width=\textwidth]{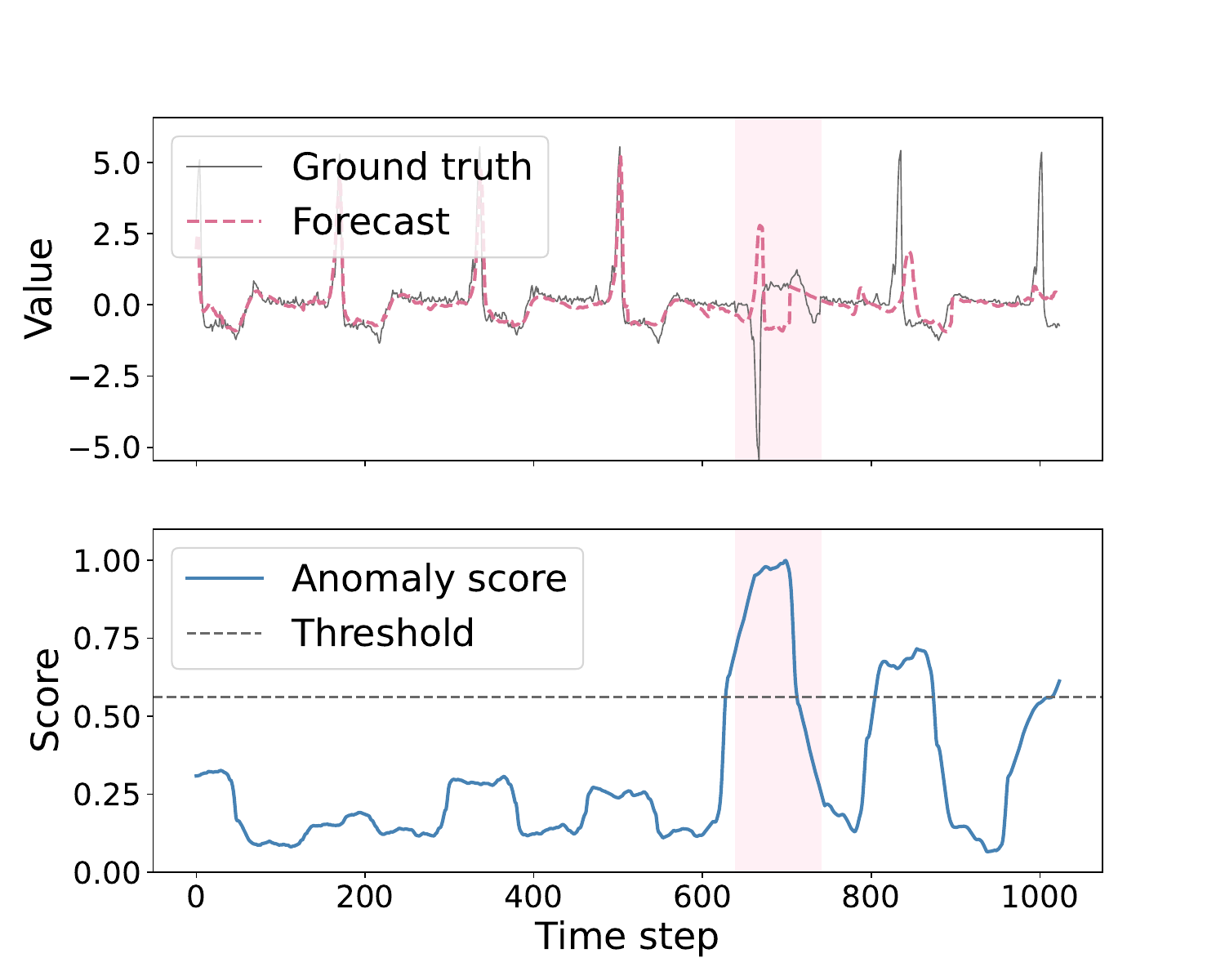}
            \caption*{\textbf{RATFM}.}
        \end{subfigure}
        \hspace{0.02\textwidth}
        \begin{subfigure}{0.33\textwidth}
            \centering
            \includegraphics[width=\textwidth]{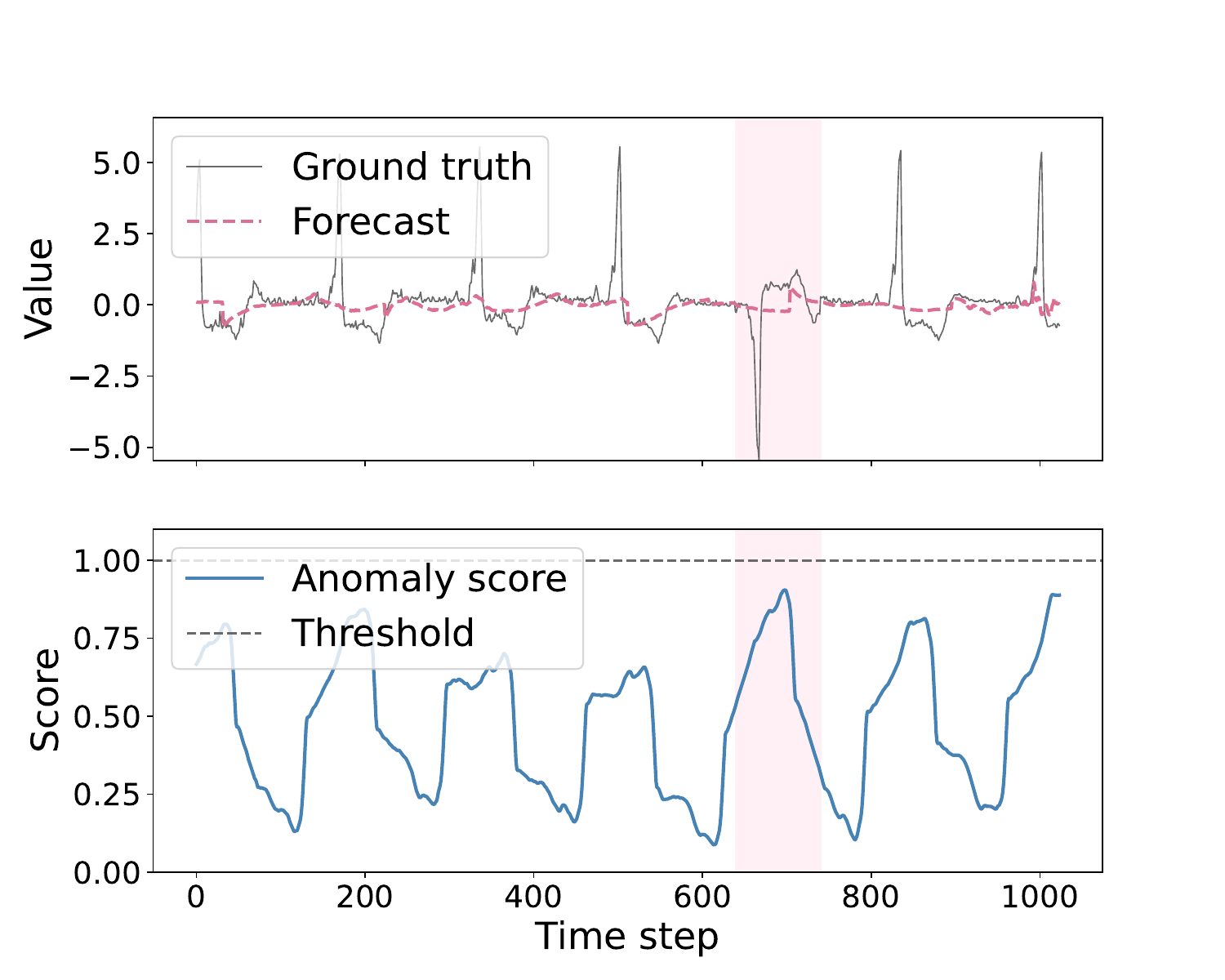}
            \caption*{\textbf{RATFM w/o training}.}
        \end{subfigure}

        \caption{\textbf{Time-MoE}.}
    \end{subfigure}

    \vspace{1em}

    \begin{subfigure}{\textwidth}
        \centering
        \begin{subfigure}{0.33\textwidth}
            \centering
            \includegraphics[width=\textwidth]{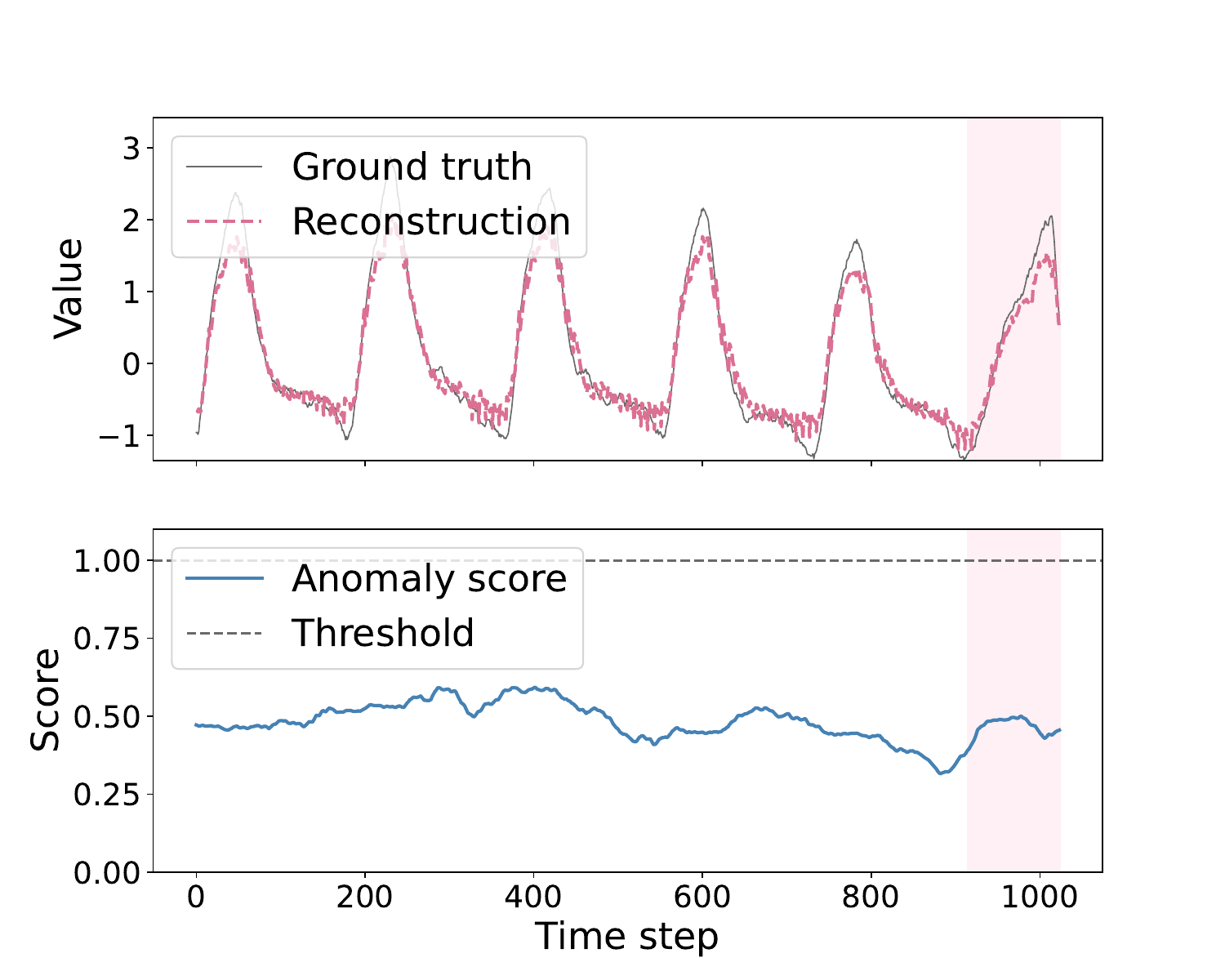}
            \caption*{\textbf{Zero-shot}.}
        \end{subfigure}
        \hspace{-0.01\textwidth}
        \begin{subfigure}{0.33\textwidth}
            \centering
            \includegraphics[width=\textwidth]{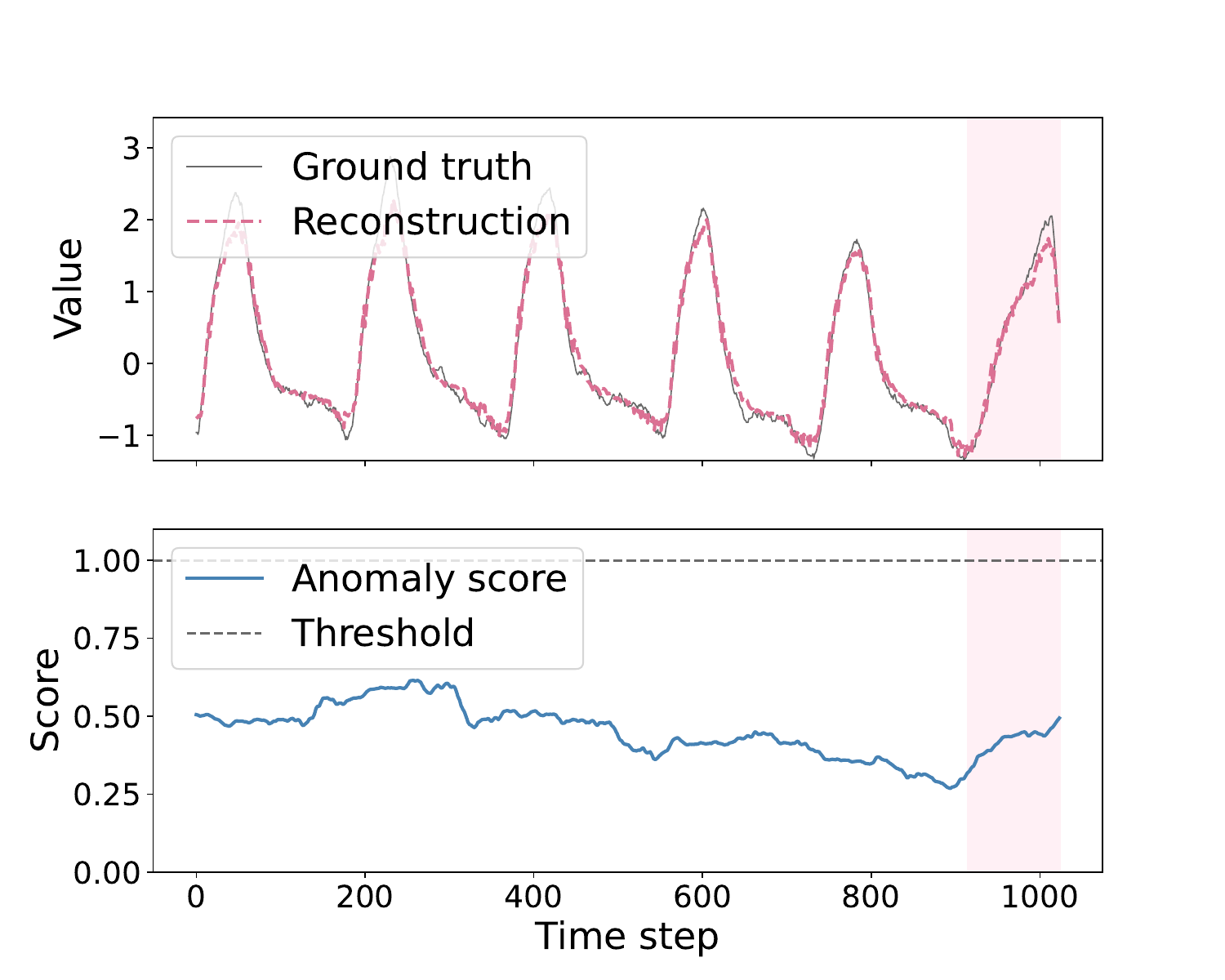}
            \caption*{\textbf{Out-domain FT}.}
        \end{subfigure}
        \hspace{-0.01\textwidth}
        \begin{subfigure}{0.33\textwidth}
            \centering
            \includegraphics[width=\textwidth]{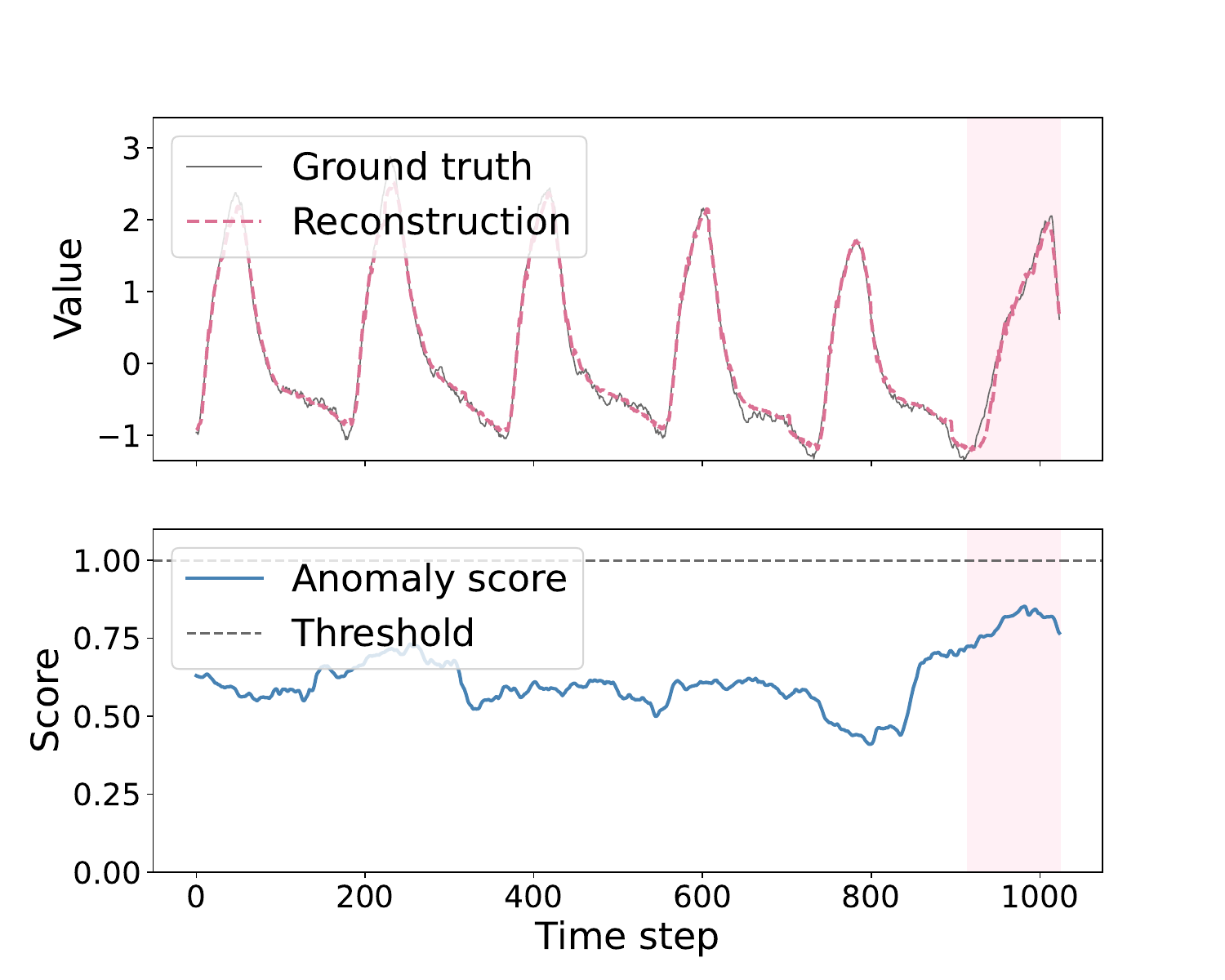}
            \caption*{\textbf{In-domain FT}.}
        \end{subfigure}

        \vspace{0.3em}

        \begin{subfigure}{0.33\textwidth}
            \centering
            \includegraphics[width=\textwidth]{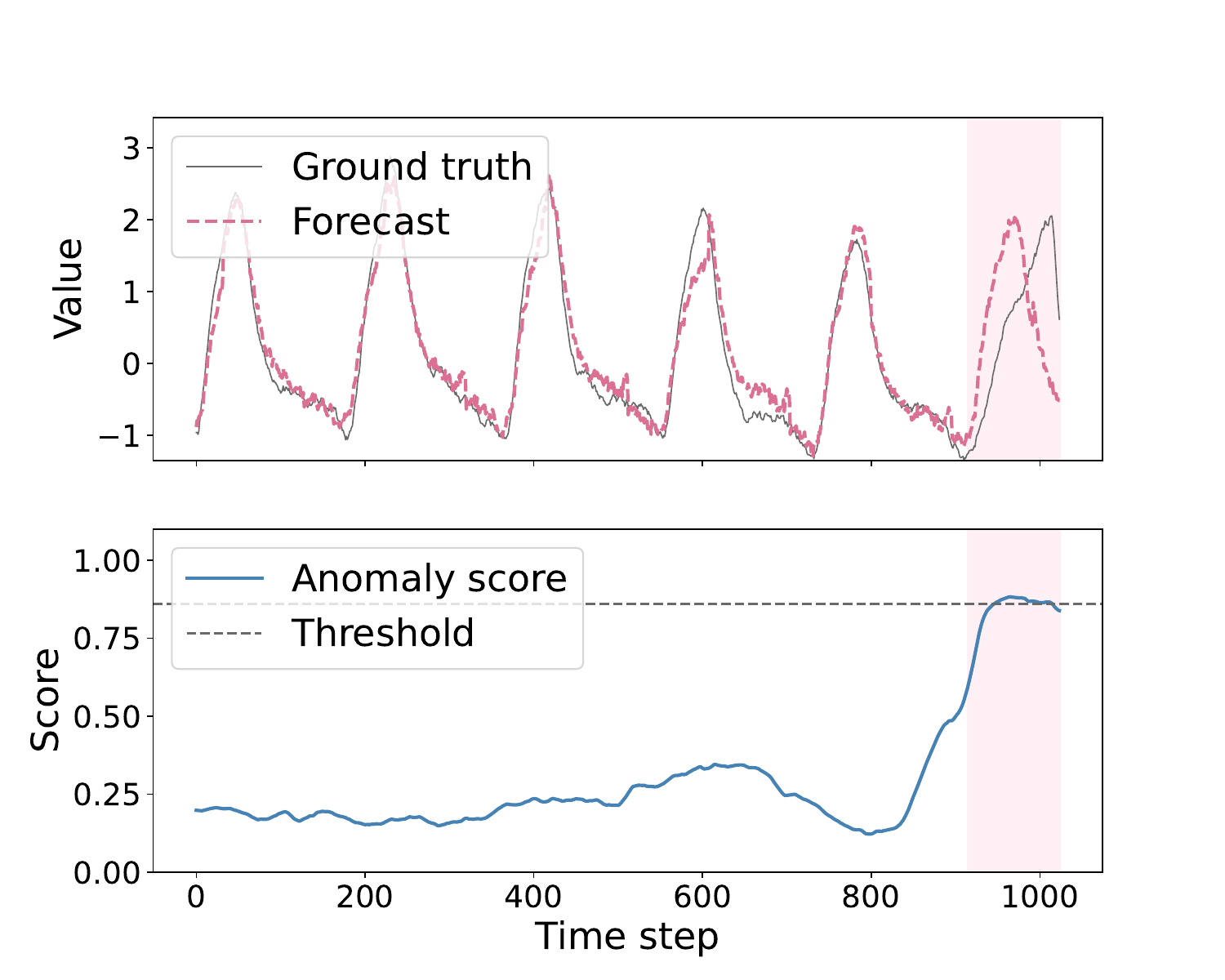}
            \caption*{\textbf{RATFM (forecast)}.}
        \end{subfigure}
        \hspace{-0.01\textwidth}
        \begin{subfigure}{0.33\textwidth}
            \centering
            \includegraphics[width=\textwidth]{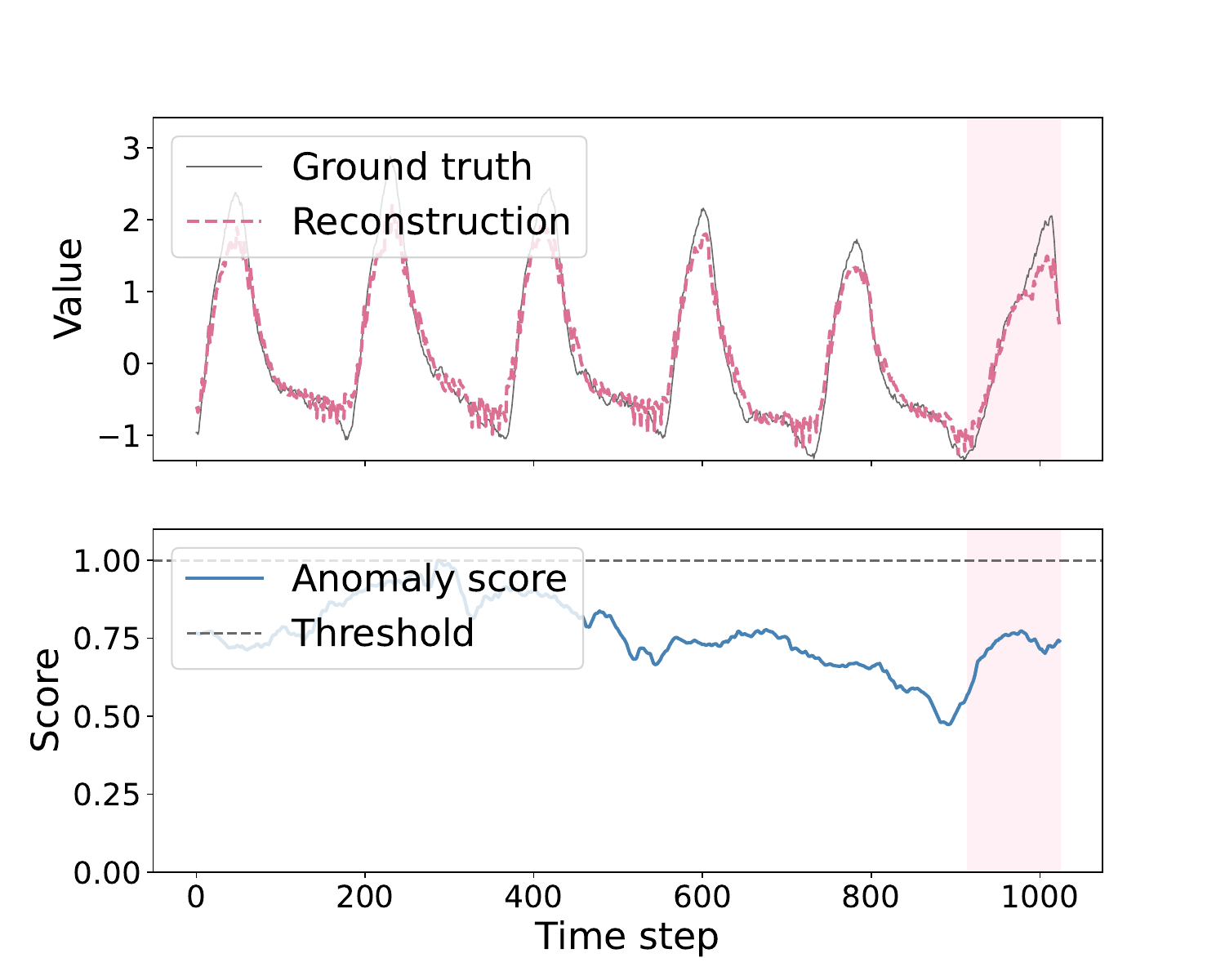}
            \caption*{\textbf{RATFM w/o training}.}
        \end{subfigure}
        \hspace{-0.01\textwidth}
        \begin{subfigure}{0.33\textwidth}
            \centering
            \includegraphics[width=\textwidth]{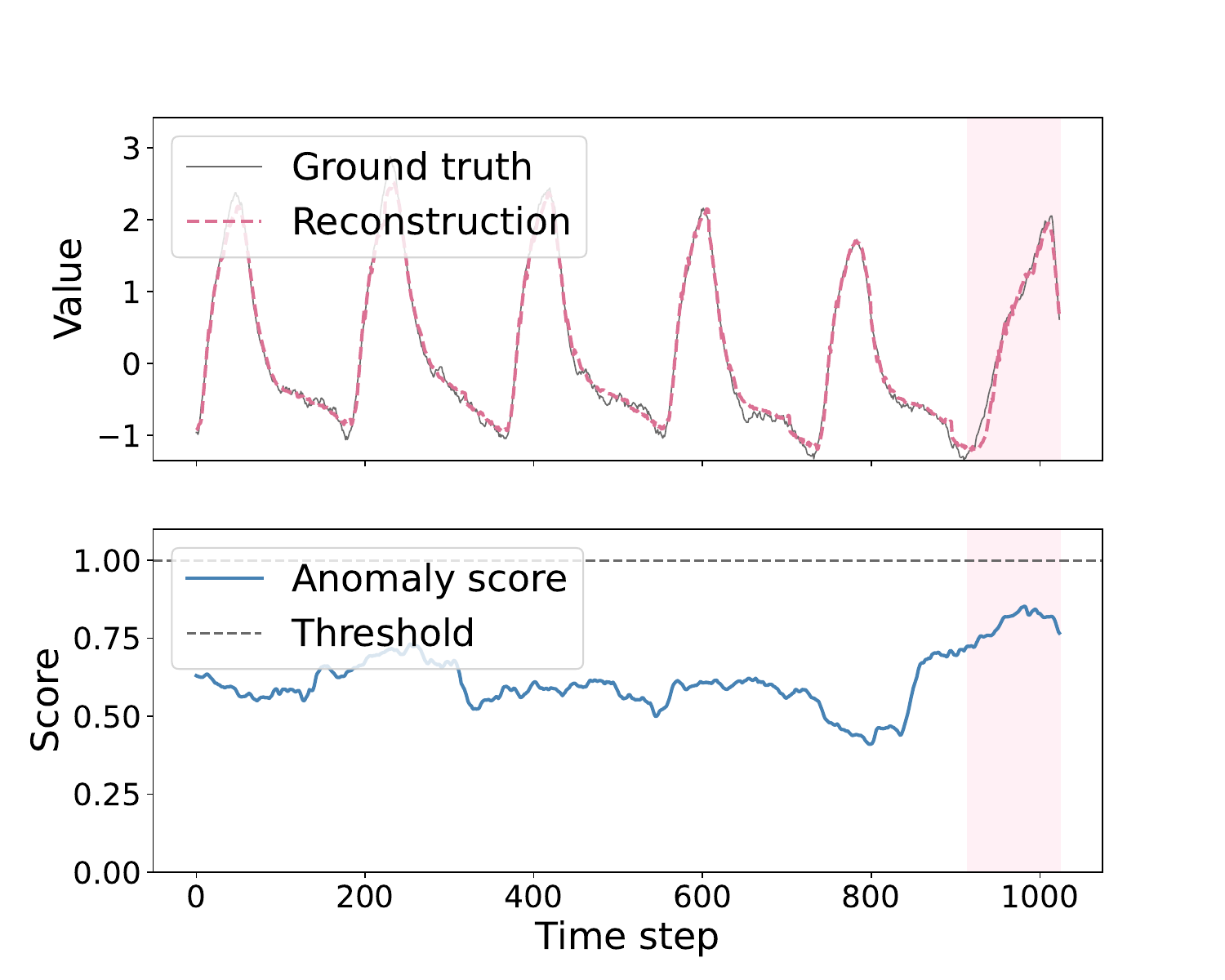}
            \caption*{\textbf{RATFM (reconst.)}.}
        \end{subfigure}

        \caption{\textbf{Moment}.}
    \end{subfigure}

    \caption{Time series forecasting results (top) and anomaly scores after applying SMA (bottom) under different settings. The highlighted regions indicate ground-truth anomalies.}
    \label{fig:main_examples_combined}
\end{figure*}

\figref{main_examples_combined} presents the time series forecasting results and anomaly scores under different settings for \textbf{Time-MoE} and \textbf{Moment}. The highlighted regions indicate the ground-truth anomalies. 
In the \textbf{RATFM} setting, the anomaly scores were high within the highlighted regions, indicating that the anomalies were correctly detected.

\end{document}